%% file: main.tex
\documentclass[11pt, a4paper, logo, copyright, nonumbering]{map}
\input{resources/packages}

\definecolor{hidden-draw}{RGB}{20,68,106}
\definecolor{hidden-pink}{RGB}{255,245,247}
\usepackage[edges]{forest}

\usepackage{bm}

\usepackage{natbib}
\usepackage{CJKutf8}
\usepackage{xargs}
\usepackage{todonotes}
\usepackage{multirow}
\usepackage{cleveref}
\usepackage{amsmath}
\usepackage{dsfont}
\usepackage{subcaption}
\usepackage{url}
\usepackage{svg}
\usepackage{fontawesome}
\usepackage{xcolor}
\usepackage{float}

\usepackage[utf8]{inputenc}
\usepackage{booktabs}
\usepackage{graphicx}
\usepackage{array}
\usepackage{tabularx}
\usepackage{xcolor,colortbl}  
\definecolor{boxcolor}{HTML}{d92523} 
\definecolor{bulbcolor}{HTML}{e3b87f} 
\usepackage{url}
\usepackage{ragged2e}   
\usepackage{makecell} 

\usepackage{hyperref}       
\usepackage{url}            
\usepackage{booktabs}       
\usepackage{amsfonts}       
\usepackage{nicefrac}       
\usepackage{microtype}      
\usepackage{xcolor}         
\usepackage{graphicx}
\usepackage{longtable}
\usepackage{array}
\usepackage{pbox}
\usepackage{amsmath}
\usepackage{amssymb}
\usepackage{tabularx}
\usepackage{multirow}
\usepackage{wrapfig}
\usepackage{pifont}
\usepackage{algorithm}
\usepackage{algorithmic}
\usepackage{lipsum}
\usepackage{subcaption}
\usepackage{enumitem}
\usepackage{tikz}
\usepackage{xstring}

\usepackage{color-edits}

\usepackage{booktabs}
\usepackage{multirow}
\usepackage[table]{xcolor}
\usepackage{tabularx}

\usepackage{parskip} 

\usepackage{threeparttable} 

\definecolor{rliableolive}{HTML}{BBCC33}
\definecolor{rliableblue}{HTML}{77AADD}
\definecolor{rliablered}{HTML}{f63c44}

\definecolor{rliableolive}{HTML}{BBCC33}
\definecolor{rliableblue}{HTML}{77AADD}
\definecolor{rliablered}{HTML}{f63c44}

\usepackage[most,skins,theorems]{tcolorbox}
\tcbuselibrary{skins,breakable,fitting,hooks}  
\tcbset{
  aibox/.style={
    width=\linewidth,
    top=7pt,
    bottom=2pt,
    colback=rliablered!18!white,
    colframe=black,
    colbacktitle=black,
    enhanced,
    center,
    attach boxed title to top left={yshift=-0.1in,xshift=0.15in},
    boxed title style={boxrule=0pt,colframe=white,},
  }
}
\tcbset{
  aibox2/.style={
    width=\linewidth,
    top=7pt,
    bottom=2pt,
    colback=green!18!white,
    colframe=black,
    colbacktitle=black,
    enhanced,
    center,
    attach boxed title to top left={yshift=-0.1in,xshift=0.15in},
    boxed title style={boxrule=0pt,colframe=white,},
  }
}
\newtcolorbox{AIbox}[2][]{aibox,title=#2,#1}
\newtcolorbox{AIbox2}[2][]{aibox2,title=#2,#1}

\hypersetup{
    colorlinks=true,            
    linkcolor=blue,             
    filecolor=magenta,          
    urlcolor=cyan,              
    citecolor=purple,             
    pdftitle={Overleaf Example},
    pdfpagemode=FullScreen,
}

\definecolor{iquestblue}{HTML}{173C7F}
\definecolor{iquestazure}{HTML}{528FCC}

\newcommandx{\info}[2][1=]{\todo[linecolor=red,backgroundcolor=red!25,bordercolor=red,#1]{#2}}

\usepackage{tcolorbox}
\newtcolorbox{insightbox}{
    colback=blue!5!white, 
    colframe=blue!75!black, 
    left=4pt, right=4pt, top=4pt, bottom=4pt, 
    boxrule=1pt, 
    arc=3pt, 
    width=\columnwidth 
}

\title{
\vspace{-0.2in}
\centering \fontsize{15pt}{16pt}\selectfont
Every Coin Has Two Sides: On the Dual Nature of Generalization in On-Policy Distillation of Large Language Models
\vspace{-0.2in}
}

\author{
Zhaoyi Li$^{1,3,*}$,
Deyang Kong$^{2,3,*}$,
Yuan Wei$^{3,*}$,
Evan Yang$^{3}$,
Ranran Shen$^{1}$,\\
Mahardika Krisna Ihsani$^{4}$,
Ming Yang$^{3}$,
Wei Zhang$^{3}$,
Chuan Hao$^{3}$,
Jian Yang$^{3}$,\\
Ran Tao$^{3}$,
Bryan Dai$^{3}$,
Shikun Zhang$^{2}$,
Wei Ye$^{2}$,
Ying Wei$^{5}$,
Defu Lian$^{1}$\\[3pt]
\mapaffiliation{
$^{1}$University of Science and Technology of China,
$^{2}$Peking University,\\
$^{3}$IQuest Research,
$^{4}$MBZUAI,
$^{5}$Zhejiang University
}\\
\mapaffiliation{$^{*}$Equal Contribution}\\
\mapemail{
lizhaoyi777@mail.ustc.edu.cn,
kong.deyang@foxmail.com
}
}

\input{sections/abstract}

\begin{document}

\maketitle

\let\oldthefootnote\thefootnote


\section{Introduction}
\input{sections/intro}

\input{sections/preliminary}

\input{sections/findings_1}

\input{sections/findings23}
\input{sections/discussion}
\section{Conclusion}
We study how well OPD generalizes by varying the training and evaluation distributions in a controlled way, from in-domain shifts to cross-domain transfer. OPD is largely insensitive to training problem difficulty, and it transfers a teacher's ability beyond the domain of its training prompts, but mainly for same-origin pairs. Because routing does not keep a teacher's influence within its assigned domain, combining domain experts in MOPD yields a seesaw among their capabilities rather than an isolated composition of expert skills, offering a useful perspective for diagnosing MOPD.
\section*{Limitations}

Our experiments focus on reasoning-oriented models and cover Math, Code, Science, and instruction-following domains. The observed generalization patterns may not directly extend to multimodal, tool-using, or interactive agent settings.
Extending the analysis to such tasks would help determine whether OPD exhibits similar cross-domain behavior when supervision depends on external observations or actions.
In addition, our MOPD experiments consider two teachers with complementary capabilities and use fixed domain-based prompt routing.
This controlled setting makes the cross-domain effects of individual teachers easier to analyze, but practical systems may involve larger and more heterogeneous expert pools.
They may also use adaptive router and non-uniform teacher sampling. A broader study of expert-pool size and routing strategies would provide a more complete understanding of generalization in MOPD.

\section*{Ethical Considerations}
This work studies capability transfer among publicly available language models on Math, Code, Science, and instruction-following benchmarks. Our results show that OPD can transfer teacher behavior beyond the domain of the routed training prompts. While this enables broad capability transfer, it may also propagate undesirable behaviors or biases from the teacher to domains that are not explicitly monitored during training.
In particular, prompt routing in MOPD should not be treated as a strict capability or safety boundary. Models trained with OPD or MOPD should therefore be evaluated for safety and reliability across all domains, rather than only on the domain assigned to each teacher.


\bibliography{ref}

\input{appendix/appendix}

\end{document}

%% file: resources/packages.tex
\usepackage{CJKutf8}
\usepackage{xargs}  
\usepackage{todonotes}
\usepackage{amsmath}
\usepackage{dsfont}

\usepackage{longtable}

\usepackage{svg}
\usepackage{fontawesome}
\usepackage{array}
\usepackage{tabularx}
\usepackage{latexsym}
\usepackage{graphicx}
\usepackage[utf8]{inputenc} 
\usepackage[utf8]{inputenc} 
\usepackage{amssymb} 
\usepackage{url}            
\usepackage{amsfonts}       
\usepackage{nicefrac}       
\usepackage{microtype}      
\usepackage{xspace}

\usepackage{listings}
\usepackage{makecell}
\usepackage{inconsolata}
\usepackage{multicol}
\usepackage{amstext}

\definecolor{darkblue}{RGB}{84, 112, 198}

\definecolor{lightblue}{rgb}{0.85, 0.95, 1.0}    
\definecolor{lightgreen}{rgb}{0.90, 1.0, 0.90}    
\definecolor{lightorange}{rgb}{1.0, 0.95, 0.85}   
\definecolor{lightpurple}{rgb}{0.95, 0.90, 1.0}   
\definecolor{lightgray}{rgb}{0.97, 0.97, 0.97}    

\usepackage{tikz}
\usetikzlibrary{calc}
\definecolor{battery-empty}{rgb}{0.9, 0.9, 0.9}
\newcommand{\difficultybar}[1]{%
  \begin{tikzpicture}[baseline, scale=0.5, every node/.style={scale=0.8}]
    \foreach \i in {1,2,3,4,5} {
      \ifnum\i>#1
        \draw[fill=battery-empty] (\i*0.5-0.5, 0) rectangle (\i*0.5, 0.25);
      \else
        \pgfmathsetmacro{\colorlevel}{80 - 12*(\i)} 
        \edef\x{\noexpand\draw[fill=blue!\colorlevel!white, opacity=0.9] (\i*0.5-0.5, 0) rectangle (\i*0.5, 0.25);}
        \x
        \draw[blue!50!black] (\i*0.5-0.5, 0) rectangle (\i*0.5, 0.25);
      \fi
    }
    \fill[battery-empty!70] (2.5, 0.08) rectangle (2.6, 0.17);
    \draw[battery-empty!70!black] (2.5, 0.08) rectangle (2.6, 0.17);
  \end{tikzpicture}%
}

%% file: sections/abstract.tex
\begin{abstract}
On-policy distillation (OPD) transfers teacher capabilities by supervising trajectories sampled from the student’s own policy, yet its generalization behavior remains poorly understood, as most studies evaluate OPD on a single domain and on benchmarks close to the training data. We present a controlled study that varies one generalization factor at a time, from in-domain distribution shifts to cross-domain transfer and the multi-teacher setting. We find that OPD transfers a teacher's reasoning behavior rather than its answers to particular problems: training difficulty barely matters, and even problems the teacher never solves are useful. Transfer depends strongly on the origin relationship between teacher and student: same-origin pairs bring the student close to the teacher across languages, reasoning horizons, and even other domains, whereas cross-origin pairs mostly fit the trained distribution. 
This broad reach is a double-edged sword: since routing prompts to domain experts cannot confine each teacher's influence, combining them yields a mixture-dependent \emph{seesaw} among their capabilities. These results clarify when OPD generalizes and offer a useful perspective for diagnosing multi-teacher OPD.
\end{abstract}

%% file: sections/intro.tex
On-policy distillation (OPD)~\citep{agarwal2024policy} has become increasingly important for transferring capabilities from strong teacher models to smaller or less capable students during large language model post-training~\cite{coreteam2026mimov2flashtechnicalreport,kimiteam2026kimik3openfrontier,glm5team2026glm5vibecodingagentic}. Unlike conventional offline distillation, OPD samples trajectories from the student’s current policy and queries the teacher on states that the student actually visits. The resulting supervision therefore reduces the exposure bias~\citep{lu2025onpolicydistillation} between the training trajectories and the student’s own generation behavior.
Most existing studies evaluate OPD in a single training domain and on benchmarks closely related to the training data.
Such evaluations establish that OPD can improve target-task performance, but they do not distinguish \emph{local fitting} from \emph{broader policy transfer}~\citep{chu2025sft}, a distinction critical for both understanding the mechanisms of OPD~\citep{li2026rethinking} and designing effective multi-teacher integration strategies~\citep{ma2026mopd}. 
In particular, it remains unclear how OPD behaves when training and evaluation differ in language, reasoning horizon or task domain.

To fill this gap, we conduct a controlled study that varies one generalization factor at a time while holding the remaining conditions fixed, and organize it around three questions of increasing scope.
\textbf{\emph{RQ1: How robust is OPD to in-domain distribution shifts?}}
Within a fixed domain (math), we study two aspects. First, we examine how the relative difficulty~\citep{chandra2025shape,yu2026dapo,zheng2026scope} of the training problems, measured by the teacher's and the student's pass rates, affects the transfer of the teacher's in-domain performance~\citep{zheng2026scope,hou2026uni}. Second, keeping the domain fixed but shifting the evaluation distribution, in language~\citep{liu2025livemathbench,sun2026olymmath} (English$\rightarrow$Chinese) and reasoning horizon~\citep{lu2025r,shojaee2025the} (short$\rightarrow$long-horizon composed problems), we examine how well the teacher's performance transfers under such shifts.
For \textbf{RQ1}, OPD is largely insensitive to training-problem difficulty: problems the teacher never solves are as useful as those it always solves, suggesting OPD conveys the teacher's reasoning patterns rather than answers to particular problems.
The transferred ability also holds up under the evaluation shifts: training only on English short-horizon math still improves the student on Chinese and long-horizon math.
\textbf{\emph{RQ2: To what extent does OPD transfer across domains?}}
We ask whether supervision from math prompts improves the student on code and science, whether prompts from other domains transfer back to math, and how model origin shapes this transfer~\citep{luo2026demystifying}.
For \textbf{RQ2}, OPD transfers a teacher's ability beyond the domain of its training prompts in both directions, but this transfer holds mainly for \emph{same-origin} pairs, which bring the student close to the teacher's level across domains, whereas \emph{cross-origin} pairs improve the student mainly on the trained distribution and can transfer less than a weaker same-origin teacher.
\textbf{\emph{RQ3: What does cross-domain transfer imply for multi-teacher OPD (MOPD)?}}
MOPD~\citep{ma2026mopd} routes each prompt to a domain expert, seemingly isolating their contributions. We ask whether this holds, given that each teacher's influence extends beyond its assigned domain.
For \textbf{RQ3}, this same transfer has a cost: because routing does not keep a teacher's influence within its assigned domain, combining experts in MOPD produces a mixture-dependent \emph{seesaw} among their capabilities rather than independently combining domain skills.
The generalization that is a blessing in single-teacher OPD is thus also what makes MOPD hard to control, two sides of the same coin, and the seesaw offers a useful perspective for diagnosing MOPD.

Taken together, our findings suggest a unified picture. (1) OPD transfers the teacher’s reasoning behavior rather than solutions to particular problems: in domain, training difficulty barely matters, and even problems the teacher never solves are as useful as those it always does. (2) The reach of this transfer is affected by the origin relationship between teacher and student. A same-origin teacher moves the student close to its own level across languages, horizons, and even other domains, whereas a cross-origin teacher affects the student mainly on the trained distribution and can transfer less than a weaker same-origin teacher. (3) This broad transfer is a double-edged sword: because each teacher influences capabilities beyond its assigned domain, prompt routing cannot fully isolate teacher effects in MOPD. Combining experts therefore produces a mixture-dependent capability seesaw, offering a useful perspective for diagnosing MOPD.

%% file: sections/preliminary.tex
\section{Preliminaries and Experiment Settings}
In this section, we discuss some preliminaries and basic experimental settings of this work. Please see  Appendix~\ref{appendix:related_work} for the full version of the related work.

\paragraph{OPD and MOPD}
Let $x$ denote a prompt sampled from the training dataset $\mathcal{D}$,
$\pi_{\theta}$ and
$\pi_{\phi}$ the student and teacher policies,
$y=\{y_t\}^T_{t=1}$ a student-generated response, and $h_t=(x,y_{<t})$ the context at step $t$.
On-Policy Distillation (\textbf{OPD}) samples \textbf{on-policy} trajectories from the student policy $\pi_{\theta}$ and obtains dense token-level supervision from the teacher policy $\pi_{\phi}$.
The student is optimized by minimizing $\mathcal{L}_{\mathrm{OPD}}(\theta)$, the reverse KL divergence between the student and teacher~\citep{gu2024minillm}:

{
\begin{align*}
\mathop{\mathbb{E}}\limits_{\substack{
    x\sim\mathcal{D} \\
    y\sim\pi_{\theta}(\cdot\mid x)
}}
\left[
    \frac{1}{T}
    \sum_{t=1}^{T}
    D_{\mathrm{KL}}
    \left(
        \pi_{\theta}(\cdot\mid h_t)
        \,\Vert\,
        \pi_{\phi}(\cdot\mid h_t)
    \right)
\right].
\end{align*}
}
In practice~\cite{coreteam2026mimov2flashtechnicalreport,glm5team2026glm5vibecodingagentic, kimiteam2026kimik3openfrontier}, the reverse KL is estimated by a sampled-token $k_1$ approximation~\citep{joschuApproximatingDivergence}, so that
$\mathcal{L}_{\mathrm{OPD}}(\theta)$ can be written as a reinforcement-learning objective (i.e., \textbf{Policy-Gradient (PG) style OPD}):

{
\begin{align*}
&\mathcal{L}^{\mathrm{PG}}_{\mathrm{OPD}}(\theta)
=
-
\mathbb{E}_{x,y}
\left[
    \frac{1}{T}
    \sum_{t=1}^{T}
    \bar{A}^{\mathrm{OPD}}_t
    \log\pi_{\theta}(y_t\mid h_t)
\right],
\ \ \widehat{A}^{\mathrm{OPD}}_t
=
\operatorname{sg}
\left[
    \log\pi_{\phi}(y_t\mid h_t)
    -
    \log\pi_{\theta}(y_t\mid h_t)
\right],
\end{align*}
}
where $\operatorname{sg}[\cdot]$ denotes the stop-gradient. We adopt PG-style OPD throughout: a token is reinforced when the teacher assigns it higher probability than the student, and suppressed otherwise.
Multi-Teacher On-Policy Distillation (\textbf{MOPD})~\citep{ma2026mopd} is a natural extension of OPD and now a standard post-training paradigm~\citep{coreteam2026mimov2flashtechnicalreport,kimiteam2026kimik3openfrontier,xu2026deepseek} for integrating capabilities from multiple domains: the student samples from its own rollouts, \textbf{each prompt is routed to its corresponding domain teacher}, and the optimization procedure is identical to single-teacher OPD.

\paragraph{Same/Cross-Origin OPD}
Following~\citet{ma2026mopd}, we call an OPD run \textbf{same-origin} when the teacher and student derive from the same base model (typically the student is an SFT checkpoint and the teacher is obtained through RL post-training on that same checkpoint) and \textbf{cross-origin} when they derive from different base models. Our experiment shows that the two settings exhibit significantly different generalization behaviors.
Our teachers include \textbf{Qwen3-32B}~\citep{yang2025qwen3}, \textbf{Light-R1-14B}~\citep{wen2025lightr1}, \textbf{Polaris-7B/4B}~\citep{Polaris2025}, \textbf{OpenMath-Nemotron-1.5B/7B}~\cite{moshkov2025aimo2winningsolutionbuildingopenmath}, JustRL-DeepSeek-1.5B (\textbf{JustRL-1.5B})~\citep{he2025justrl}, Nemotron-Research-Reasoning-Qwen-1.5B (\textbf{Nemotron-1.5B})~\citep{liu2026prorl}, \textbf{DeepScaleR-1.5B}~\cite{tan2026deepscaler} and \textbf{VibeThinker-1.5B}~\cite{xu2025tinymodelbiglogicvibethinker}, and our students include DeepSeek-R1-Distill-Qwen-14B/7B/1.5B (\textbf{DS-distill-14B/7B/1.5B})~\citep{guo2025deepseekr1}, \textbf{Qwen3-8B-SFT}~\citep{lu2025onpolicydistillation} and \textbf{Qwen3-4B}~\citep{yang2025qwen3}\footnote{Please refer to Table~\ref{tab:model_lineage} for models' post-training lineage.}. Due to the page limit, we only show part of the results in the main text. Additional results are shown in Appendix~\ref{sec:appendix_add_results}.

\paragraph{In/Cross-Domain Generalization}
We investigate OPD generalization from two aspects.
(1) \textbf{In-domain generalization:}
OPD training and evaluation lie in the same task domain, and we shift the evaluation distribution relative to training in terms of \textbf{\emph{problem difficulty}}, \textbf{\emph{language}}, and \textbf{\emph{reasoning horizon}}.
(2) \textbf{Cross-domain generalization:} the student is trained on prompts from one domain (e.g., math) and evaluated on other domains (e.g., code or science).
We study four domains: math, code, science, and instruction following (IF).
For \textbf{\underline{math}}, we train on BigMath by default and evaluate three distributions: \emph{English Math} (AMC2023~\citep{MAA_AMC}, MATH-500~\citep{hendrycks2021measuringmath500}, AIME2025~\citep{MAA_AIME}, AIME2026~\citep{MAA_AIME}, BeyondAIME~\citep{bytedance_seed_2025_beyondaime}, OlymMATH-Hard~\citep{sun2026olymmath}) as the primary evaluation benchmarks, \emph{Chinese Math} (OlymMATH-ZH~\citep{sun2026olymmath}, LiveMathBench-ZH~\citep{liu2025livemathbench}) for the language shift (generalizing from English to Chinese), and \emph{Long-Horizon Math} (the AIME24-Horizon-2 and AMC23-Horizon-4 subsets~\citep{lu2025r}, generalizing to problems that require more reasoning steps~\citep{shojaee2025the,li2026scaling}).
For \textbf{\underline{code}}, we train on \emph{DeepCoder-Preview-Dataset}~\citep{deepcoder2025} and evaluate on \emph{LiveCodeBench}~\citep{jain2025livecodebench}; 
for \textbf{\underline{science}}, we train on \emph{TextbookReasoning}~\citep{fan2025megascience} and \emph{SCP-116K}~\citep{lu2025scp} and evaluate on \emph{GPQA-Diamond}~\citep{rein2023gpqa}; for \textbf{\underline{IF}}, we train on \emph{Nemotron-Post-Training-IF}~\citep{bercovich2025llamanemotron} and evaluate on \emph{IF-Eval}~\citep{zhou2023instructionfollowingevaluationlargelanguage}. More information about the training and evaluation datasets is shown in Appendix~\ref{sec:appendix_datasets}

%% file: sections/findings_1.tex
\section{In-Domain Generalization}
We begin with the most basic form of generalization, where OPD training and evaluation stay in the same domain (math) but their distributions differ. We first ask whether the difficulty of the training problems, measured by the teacher's and the student's pass-rates, affects the in-domain generalization of OPD (Sec.~\ref{subsec:difficulty}). We then ask whether the ability acquired on English math problems generalizes to Chinese and longer-horizon problems, and how model origin shapes these behaviors (Sec.~\ref{subsec:lang_horizon}).

\subsection{Training-Problem Difficulty}
\label{subsec:difficulty}
\begin{figure*}[t]
    \centering
    \begin{subfigure}[b]{0.32\linewidth}
        \centering
        \includegraphics[width=\linewidth]{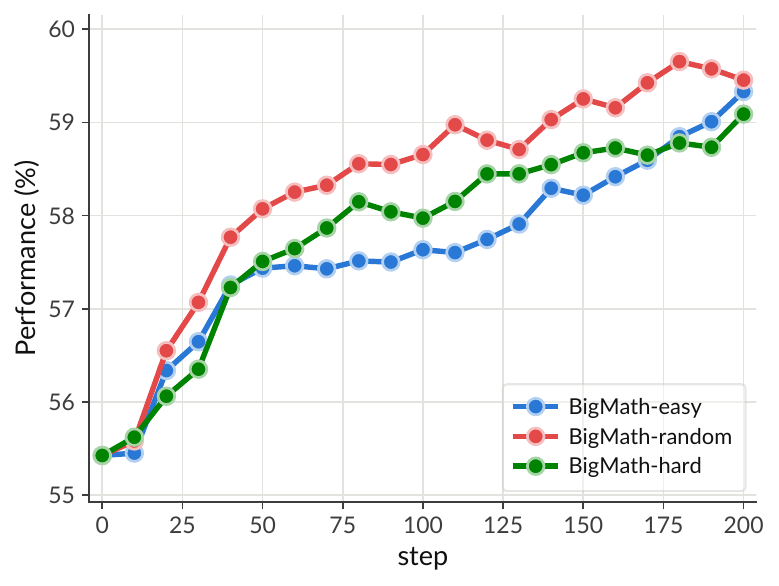}
        \caption{Qwen3-32B $\rightarrow$ Qwen3-8B-SFT}
        \label{fig:teacher_pass_32b_to_8b}
    \end{subfigure}
    \hfill
    \begin{subfigure}[b]{0.32\linewidth}
        \centering
        \includegraphics[width=\linewidth]{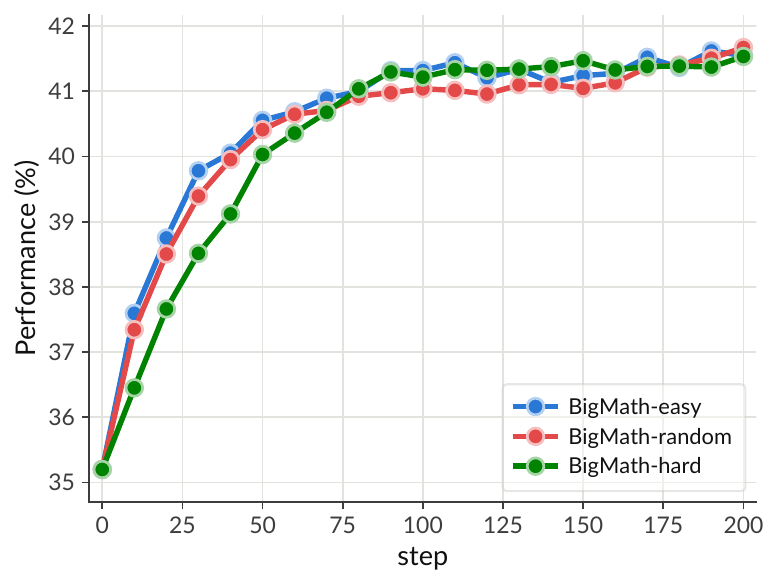}
        \caption{Polaris-7B $\rightarrow$ DS-distill-1.5B}
        \label{fig:teacher_pass_7b_to_1d5b}
    \end{subfigure}
    \hfill
    \begin{subfigure}[b]{0.32\linewidth}
        \centering
        \includegraphics[width=\linewidth]{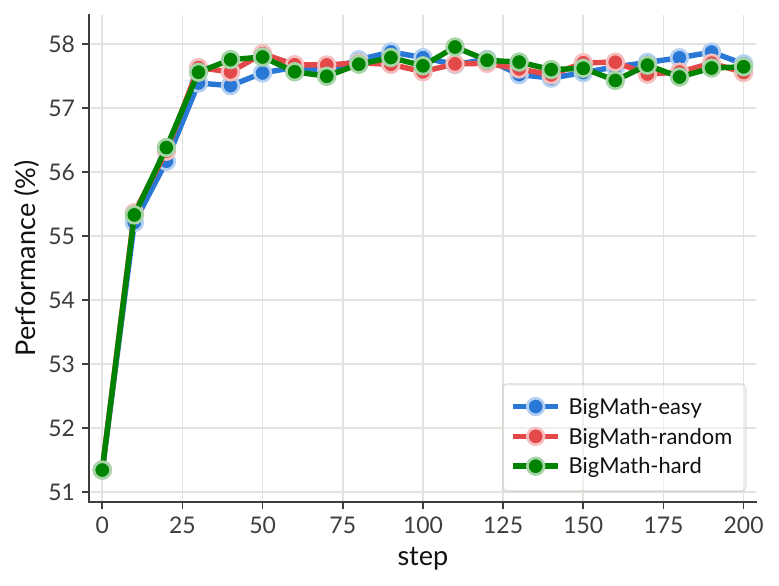}
        \caption{Polaris-7B $\rightarrow$ DS-distill-7B}
        \label{fig:teacher_pass_7b_to_7b}
    \end{subfigure}

    \caption{In-domain math accuracy (average over six English benchmarks) is insensitive to the difficulty of the training problems, measured by the teacher's pass-rate. Each figure compares three BigMath subsets: \textbf{easy} (pass-rate $=1$, teacher solves all four rollouts), \textbf{hard} (pass-rate $=0$, teacher solves none), and \textbf{random} (randomly sampled from the whole set). The three subsets converge to nearly identical final accuracy across all teacher--student pairs.}
    \label{fig:teacher_pass_rate_combined}
\end{figure*}

We measure problem difficulty from two sides: the \textbf{\emph{teacher}'s pass-rate}, a fixed difficulty estimate available before training, and the \textbf{\emph{student}'s pass-rate}, a dynamic signal that evolves during training.
\paragraph{Teacher pass-rate.}
To study how teacher-end pass-rate affects the in-domain generalization of OPD, we sample four teacher responses per BigMath problem and form three subsets of 25K problems each: \textbf{easy} (pass-rate $=1$), \textbf{hard} (pass-rate $=0$), and \textbf{random} (randomly sampled from the whole set), keeping all other conditions fixed. As Figure~\ref{fig:teacher_pass_rate_combined} shows, the three subsets converge to nearly identical final accuracy across all teacher--student pairs. This demonstrates that \emph{\textbf{OPD teaches student models the teachers' reasoning patterns rather than the correct answers to particular problems}}, so a teacher can still supply informative token-level supervision on problems it cannot solve end to end~\cite{chandra2025shape}. 
We also conduct experiments with extremely easy problems (grade-school GSM8K) and difficult problems (the hardest slice of DeepMath-103K), which gives the similar observation: training on these data still recovers over $80\%$ of the OPD gain of the default BigMath-random (Fig.~\ref{fig:extreme_difficulty} in Appendix~\ref{sec:appendix_add_results}). Teacher-side filtering by difficulty therefore makes little difference.
\begin{insightbox}
\textbf{Takeaway:}
OPD transfers a teacher's \emph{reasoning behavior}, not its answers to particular problems. Within math, the difficulty of the training problems, measured by the teacher's pass rate, has little effect on in-domain generalization: problems the teacher never solves are as useful as those it always solves, and even grade-school problems recover most of the OPD gain. 
\end{insightbox}
\begin{table*}[htbp]
\centering
\caption{Generalization performance of OPD with student-side dynamic sampling across six math benchmarks. Discarding only the problems the student already solves (pass-rate $\in[0,1)$) gives the best average performance. Green/red mark per-benchmark gains/drops relative to the w.o. dynamic sampling baseline.}
\label{tab:pass-rate-eval}
\resizebox{\textwidth}{!}{%
\begin{tabular}{c|cccccc|c}
\toprule
\textbf{Filtering Strategy} & \textbf{AMC2023} & \textbf{MATH500} & \textbf{AIME2025} & \textbf{AIME2026} & \textbf{BeyondAIME} & \textbf{OlymMATH-Hard} & \textbf{Average} \\
\midrule
\midrule
\multicolumn{8}{c}{\textbf{\textit{Teacher: Polaris-7B, Student: DS-distill-1.5B}}} \\
\midrule
\textbf{w/o dynamic sampling} & $80.4\%$ & $89.5\%$ & $30.2\%$ & $30.1\%$ & $13.6\%$ & $4.7\%$ & $41.4\%$ \\
\midrule
\textbf{pass-rate} $\mathbf{= 0}$        & \cellcolor{red!25}$78.8\%$ & \cellcolor{green!25}$90.1\%$ & \cellcolor{green!25}$30.5\%$ & \cellcolor{red!25}$30.0\%$ & \cellcolor{green!25}$14.7\%$ & \cellcolor{red!25}$4.4\%$ & $41.4\%$ \\
\textbf{pass-rate} $\mathbf{= 1}$        & \cellcolor{red!25}$79.9\%$ & \cellcolor{green!25}$89.7\%$ & \cellcolor{green!25}$30.9\%$ & \cellcolor{red!25}$28.9\%$ & \cellcolor{green!25}$14.7\%$ & \cellcolor{red!25}$4.0\%$ & $41.4\%$ \\
\textbf{pass-rate} $\mathbf{\in [0, 1)}$ & \cellcolor{green!25}$80.9\%$ & \cellcolor{green!25}$89.8\%$ & \cellcolor{green!25}$31.2\%$ & \cellcolor{green!25}$31.1\%$ & \cellcolor{green!25}$13.7\%$ & \cellcolor{green!25}$5.4\%$ & $\mathbf{42.0}\%${\footnotesize \textcolor{green!60!black}{(\textbf{+0.6 pp})}} \\
\midrule
\midrule
\multicolumn{8}{c}{\textbf{\textit{Teacher: Light-R1-14B, Student: DS-distill-7B}}} \\
\midrule
\textbf{w.o. dynamic sampling} & $91.7\%$ & $95.0\%$ & $42.8\%$ & $49.7\%$ & $26.5\%$ & $8.7\%$ & $52.4\%$ \\
\midrule
\textbf{pass-rate} $\mathbf{= 0}$        & \cellcolor{green!25}$92.1\%$ & \cellcolor{red!25}$94.3\%$ & \cellcolor{red!25}$40.2\%$ & \cellcolor{green!25}$50.3\%$ & \cellcolor{green!25}$28.2\%$ & \cellcolor{red!25}$7.8\%$ & $52.1\%${\footnotesize \textcolor{red!60!black}{(\textbf{-0.3 pp})}} \\
\textbf{pass-rate} $\mathbf{= 1}$        & \cellcolor{green!25}$92.3\%$ & \cellcolor{red!25}$94.5\%$ & \cellcolor{red!25}$42.2\%$ & \cellcolor{red!25}$48.7\%$ & \cellcolor{green!25}$28.0\%$ & \cellcolor{red!25}$8.0\%$ & $52.2\%${\footnotesize \textcolor{red!60!black}{(\textbf{-0.2 pp})}} \\
\textbf{pass-rate} $\mathbf{\in [0, 1)}$ & \cellcolor{green!25}$92.1\%$ & \cellcolor{green!25}$95.4\%$ & \cellcolor{green!25}$43.2\%$ & \cellcolor{green!25}$50.6\%$ & \cellcolor{green!25}$28.0\%$ & \cellcolor{red!25}$7.5\%$ & $\mathbf{52.8}\%${\footnotesize \textcolor{green!60!black}{(\textbf{+0.4 pp})}} \\
\bottomrule
\end{tabular}%
}
\end{table*}

\paragraph{Student pass-rate.}
We test two pairs, Polaris-7B\,$\rightarrow$\,DS-distill-1.5B and Light-R1-14B\,$\rightarrow$\,DS-distill-7B, both on the random subset of BigMath. 
For each problem we sample four student responses and compare three strategies against the no-filtering baseline: keeping only fully unsolved problems (pass-rate $=0$), only fully solved problems (pass-rate $=1$), or discarding only fully solved problems (pass-rate $\in[0,1)$). As Table~\ref{tab:pass-rate-eval} shows, restricting to either extreme does not help, whereas \emph{\textbf{dynamically discarding only the problems the student already solves gives a small but consistent gain}} for both pairs. The result is intuitive: once the student reliably solves a problem, the teacher should not keep realigning its reasoning there. In summary, OPD's in-domain generalization is largely insensitive to training-problem difficulty.
\begin{insightbox}
\textbf{Takeaway:}
Dynamically discarding problems the student already solves (i.e., pass-rate = 0) gives a small but consistent improvement, indicating that a teacher should not keep realigning reasoning the student has mastered. 
\end{insightbox}
\subsection{Generalization across Language and Reasoning Horizon}
\label{subsec:lang_horizon}

\begin{figure*}[t]
    \centering

    \begin{subfigure}[b]{0.99\linewidth}
        \centering
        \includegraphics[width=\linewidth]{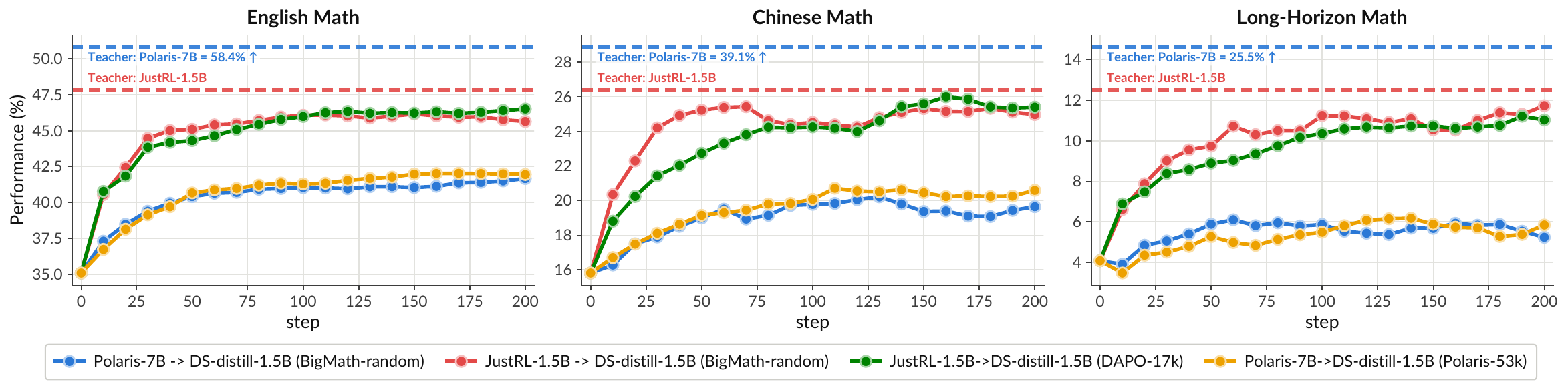}
        \caption{DS-distill-1.5B}
        \label{fig:1d5b_cross_language_horizon}
    \end{subfigure}

    \begin{subfigure}[b]{0.99\linewidth}
        \centering
        \includegraphics[width=\linewidth]{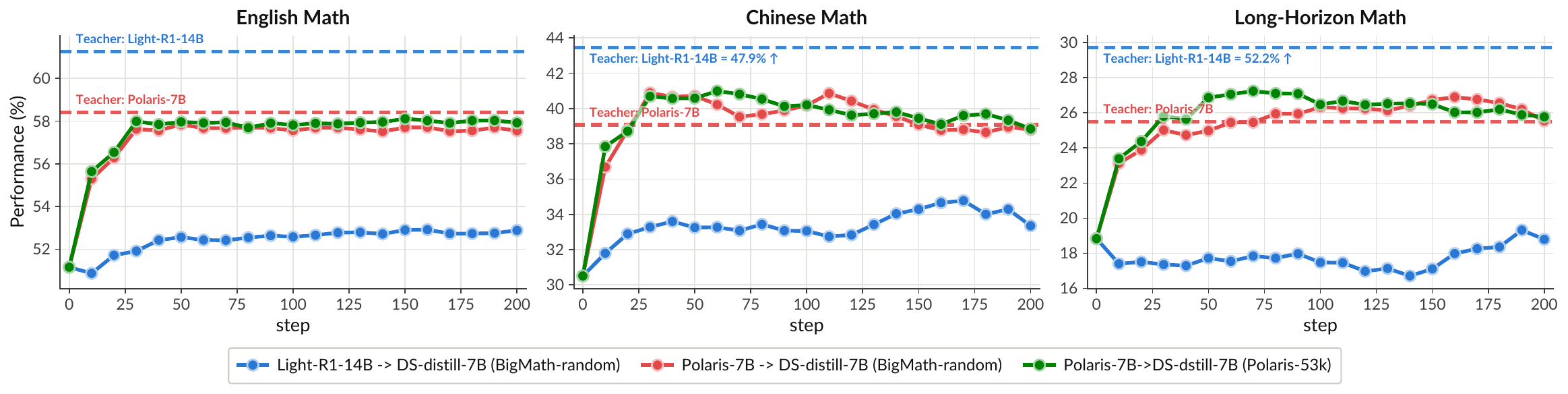}
        \caption{DS-distill-7B}
        \label{fig:7b_cross_language_horizon}
    \end{subfigure}

    \caption{OPD trained only on English math problems generalizes to Chinese and long-horizon math benchmarks, but the size and stability of the gains depend on model origin. Each figure pairs a same-origin and a cross-origin teacher for the same student; dashed lines mark the teachers' accuracy.}
    \label{fig:cross_language_and_horizon_same_different_origin}
\end{figure*}

We ask whether the acquired reasoning ability generalizes to two shifted distributions: (1) the training data are English math problems while the evaluation data are Chinese math problems; (2) the training problems are originally atomic and short-horizon, while the evaluation problems are long-horizon~\citep{dziri2023faith, shojaee2025the,lu2025r}, formed by \emph{composing multiple math problems together}; this probes systematic generalization. We evaluate two students, DS-distill-1.5B/7B, each paired with a same-origin and a cross-origin teacher, all trained on BigMath-random.
Figure~\ref{fig:cross_language_and_horizon_same_different_origin} shows that \emph{\textbf{OPD improves not only the in-distribution English math performance but also the Chinese and long-horizon distributions}}. Training on English data alone raises Chinese math accuracy, so the acquired ability is not tied to the surface language of the training problems; gains also appear on the long-horizon benchmarks, whose problems require long-horizon reasoning and propagating intermediate answers across sub-problems, a structure absent from the training data. However, the size of these gains differ markedly between the same- and cross-origin teachers, which we examine next.

\paragraph{The Role of Model Origin.}
\label{subsec:origin_indomain}
For each student, we compare OPD generalization under same-origin and cross-origin teachers (e.g., for DS-distill-1.5B, JustRL-1.5B is the same-origin teacher while Polaris-7B is the cross-origin teacher) in Figure~\ref{fig:cross_language_and_horizon_same_different_origin}.
Two observations stand out.
First, \emph{\textbf{a stronger teacher does not result in a stronger student: same-origin teachers are much more effective than cross-origin teachers}}. For instance, in the 7B panel the cross-origin Light-R1-14B has clearly higher standalone accuracy than the same-origin Polaris-7B (dashed lines), yet produces a much weaker student, and on long-horizon math the cross-origin student shows almost no significant gain over the initial model. Second, \emph{\textbf{same-origin OPD brings the student consistently close to the teacher's accuracy, and this holds not only on the in-distribution English evaluation but also under the language and horizon shifts}}. A plausible explanation is that origin serves as an indicator of how compatible the teacher's policy is with the student's, so that \emph{\textbf{same-origin pairs can align as a whole and carry that alignment across distributions}}. 
In addition, the empirical results demonstrate that using teacher's RL training prompts (e.g., Polaris-53k for Polaris-7B, and DAPO-17k for JustRL-1.5B) produces no substantial difference in OPD performance.

\begin{insightbox}
\textbf{Takeaway:}
OPD also works across in-domain distribution shifts, but mainly in the same-origin setting: trained only on English, short-horizon math, a student with a same-origin teacher approaches the teacher's level on Chinese and long-horizon math, whereas a cross-origin teacher (even with better performance than the same-origin teacher) yields smaller gains.
\end{insightbox}

%% file: sections/findings23.tex
\begin{figure*}[t]
    \centering
    \captionsetup[subfigure]{font=footnotesize}
    \begin{subfigure}[b]{0.32\linewidth}
        \centering
        \includegraphics[width=\linewidth]{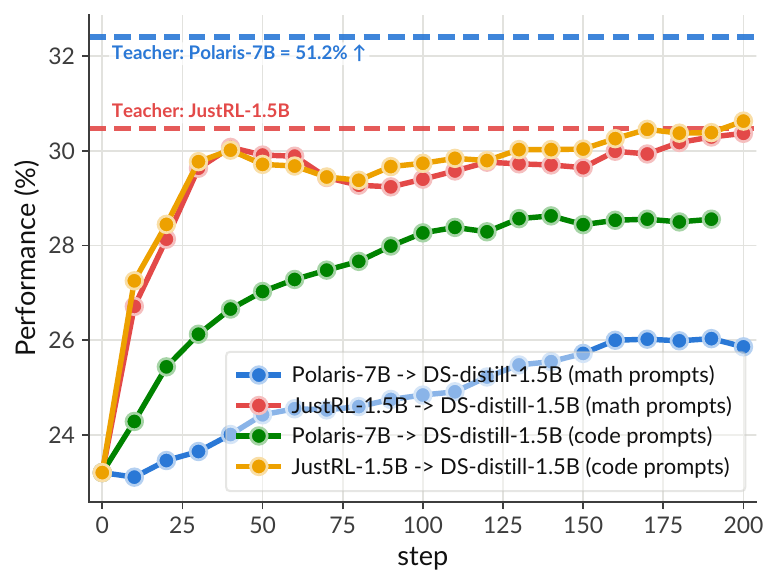}
        \caption{LiveCodeBench (DS-distill-1.5B)}
        \label{fig:math_to_code_1d5b}
    \end{subfigure}
    \hfill
    \begin{subfigure}[b]{0.32\linewidth}
        \centering
        \includegraphics[width=\linewidth]{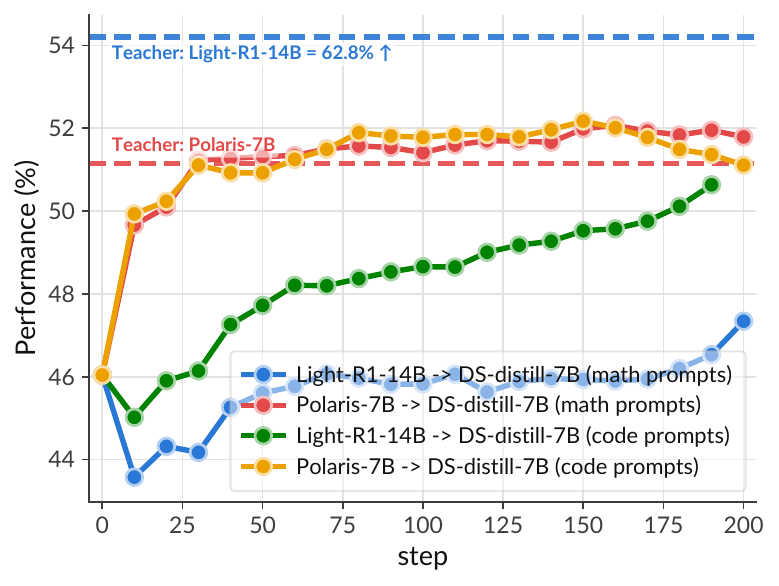}
        \caption{LiveCodeBench (DS-distill-7B)}
        \label{fig:math_to_code_7b}
    \end{subfigure}
    \hfill
    \begin{subfigure}[b]{0.32\linewidth}
        \centering
        \includegraphics[width=\linewidth]{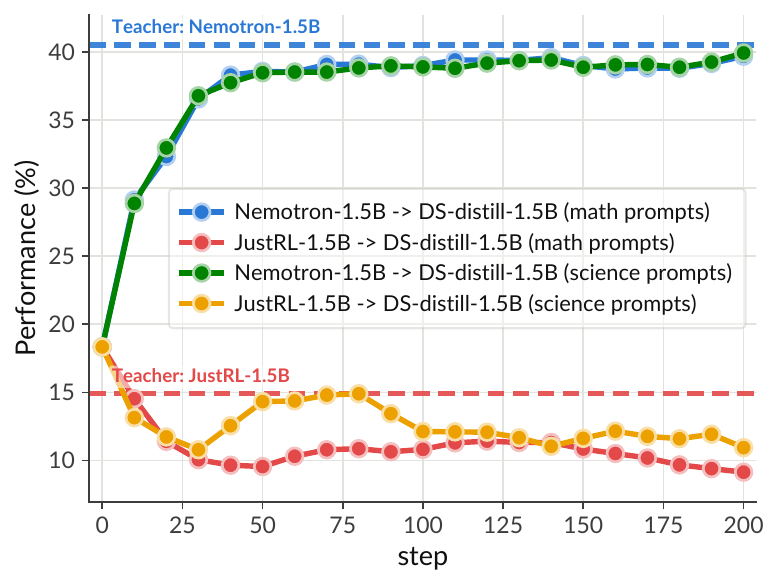}
        \caption{GPQA-Diamond (DS-distill-1.5B)}
        \label{fig:math_to_science_1d5b}
    \end{subfigure}

     \begin{subfigure}[b]{0.99\linewidth}
        \centering
        \includegraphics[width=\linewidth]{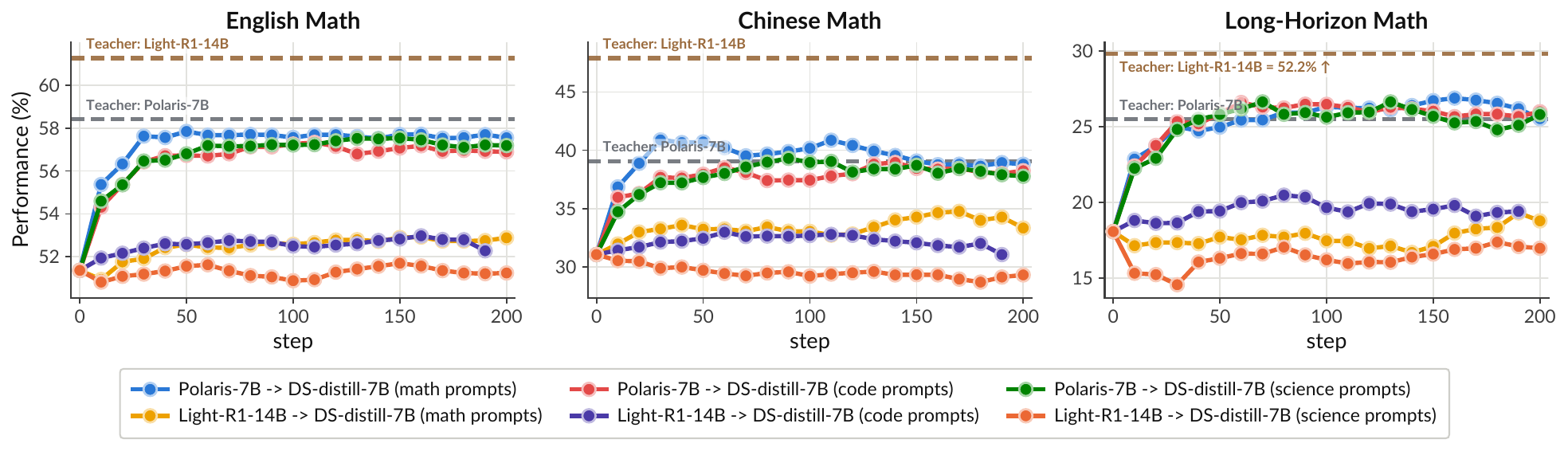}
        \caption{Training on non-math domains, evaluating on math (DS-distill-7B)}
        \label{fig:generalize_to_math_7b_panel}
    \end{subfigure}

    \caption{Cross-domain generalization in single-teacher OPD. For same-origin OPD, training on math prompts, trained student's performance on science and code also approaches teacher's performance and vice versa. For cross-origin OPD,
    cross-domain generalization is much worse: in (a) and (b) (LiveCodeBench evaluation results), the \textcolor{green!80!black}{green curves} (training on code prompts) are \emph{\textbf{consistently and substantially higher}} than the \textcolor{blue}{blue curves} (training on math prompts).
    Dashed lines mark the teachers' standalone accuracy.}
    \label{fig:cross_domain_generalization_math_to_others}
\end{figure*}
\section{Cross-Domain Generalization and Its Implication for MOPD}
\label{sec:cross_domain}
We now turn to cross-domain generalization, where OPD is performed on prompts from one domain and evaluated on other domains, and to its implications for MOPD. This section develops two connected findings. First, for the same-origin OPD setting, training on prompts from one domain also enables the student to approach the teacher's performance in \emph{other} domains. In contrast, the cross-origin setting exhibits a clear gap between cross-domain and in-domain generalization. Consequently, a same-origin teacher's influence is not confined to the domain of its training prompts, routing prompts to different domain expert teachers in MOPD does not isolate their effects; changing the teacher mixture ratios pulls the student between the domain experts, producing a \emph{seesaw effect}.
\subsection{Cross-Domain Generalization of OPD}
For cross-domain generalization, we study two students, DS-distill-1.5B and DS-distill-7B, and pair each with both a same-origin and a cross-origin teacher; for DS-distill-1.5B, for example, JustRL-1.5B and Nemotron-1.5B are same-origin teachers and Polaris-7B is a cross-origin one. We run OPD in two directions: training on math (teacher models' RL post-training domain) prompts and evaluating on other domains such as code and science, and training on other-domain prompts (code, science, and IF) and evaluating on math.
\paragraph{Math $\rightarrow$ Other Domains Transfer.}
We first train students on math prompts and evaluate on code and science. As Figures~\ref{fig:math_to_code_1d5b} and~\ref{fig:math_to_code_7b} show, both the 1.5B and 7B students improve clearly on LiveCodeBench, although no code prompts are used during OPD. A similar effect appears for science: Figure~\ref{fig:math_to_science_1d5b} compares Nemotron-1.5B and JustRL-1.5B under both math- and science-prompt training.
For the science-oriented Nemotron teacher, math-trained and science-trained runs reach similar GPQA-Diamond levels, both well above the initial student. Because the math-oriented JustRL teacher's scientific reasoning is inferior to the student's, training on either math or science prompts degrades the student's performance, ultimately driving its GPQA score below the initial point.

\paragraph{Other Domains $\rightarrow$ Math Transfer.}
As shown in Figure~\ref{fig:generalize_to_math_7b_panel}, students trained on code or science prompts improve on math (approach teacher performance), and the performance gains extend beyond English to Chinese and long-horizon math benchmarks.
These results show that a teacher's expert capability (i.e., math reasoning) can also be transferred to students even using prompts from non-primary domain (see Figure~\ref{fig:if_to_math_cross_domain_generalization} for IF$\rightarrow$math transfer).
Together, these results show that \textbf{\emph{OPD can transfer capabilities beyond the semantic domain of its training prompts: supervision collected from math reasoning can pull the student's code/scientific reasoning abilities towards the teachers, and vice versa}}.

\begin{insightbox}
\textbf{Takeaway:}
OPD can transfer a same-origin teacher's ability beyond the domain of its training prompts, in both directions: training on math prompts improves the student on code and science, and training on code, science, or instruction-following prompts transfers back to math, with gains extending to Chinese and long-horizon math.
\end{insightbox}

\paragraph{Generalization Differs by Model Origin in OPD.}
Model origin, first seen in Sec.~\ref{subsec:origin_indomain}, reappears clearly in the cross-domain setting. In Figures~\ref{fig:math_to_code_1d5b} and~\ref{fig:math_to_code_7b}, same-origin teacher runs form consistent training curves for the \textcolor{yellow!85!black}{code-prompts} training and \textcolor{red}{math-prompts} training on LiveCodeBench: both of them \textbf{approach} the teacher's performance, so within same-origin OPD the training-domain gap is largely closed and the student can absorb the teacher's code ability even from math trajectories alone. 
Cross-origin OPD behaves very differently: direct training on the target domain remains best, and we observe \textbf{a clear gap} between the in-domain generalization (training with \textcolor{green!80!black}{code prompts}) performance and cross-domain generalization (training with \textcolor{blue}{math prompts}) performance.
We attribute this difference to the distributional gap between teacher and student. For a \textit{\underline{same-origin teacher}}, whose output distribution is only slightly tuned~\citep{yue2025does} in comparison with the student's, \textbf{\emph{OPD aligns the student to the teacher at the policy level rather than only on the training domain}}. However, for a \emph{\underline{cross-origin teacher}}, the larger distributional gap makes such alignment harder, and \textbf{\textit{OPD pulls the student policy towards the teacher policy mainly on the distribution covered by the training prompts}}, leaving cross-domain generalization weaker than in-domain generalization.

\begin{insightbox}
\textbf{Takeaway:}
How far the cross-domain transfer reaches is affected by model origin. A same-origin teacher, whose distribution is close to the student's, brings the student close to the teacher's level across domains even when the prompts come from a single domain, effectively closing the training-domain gap. A cross-origin teacher, facing a larger distributional gap, improves the student mainly on the trained distribution and its cross-domain generalization stays clearly below its in-domain generalization. 
\end{insightbox}
\begin{figure*}[t]
    \centering
    \begin{subfigure}[b]{0.99\linewidth}
        \centering
        \includegraphics[width=\linewidth]{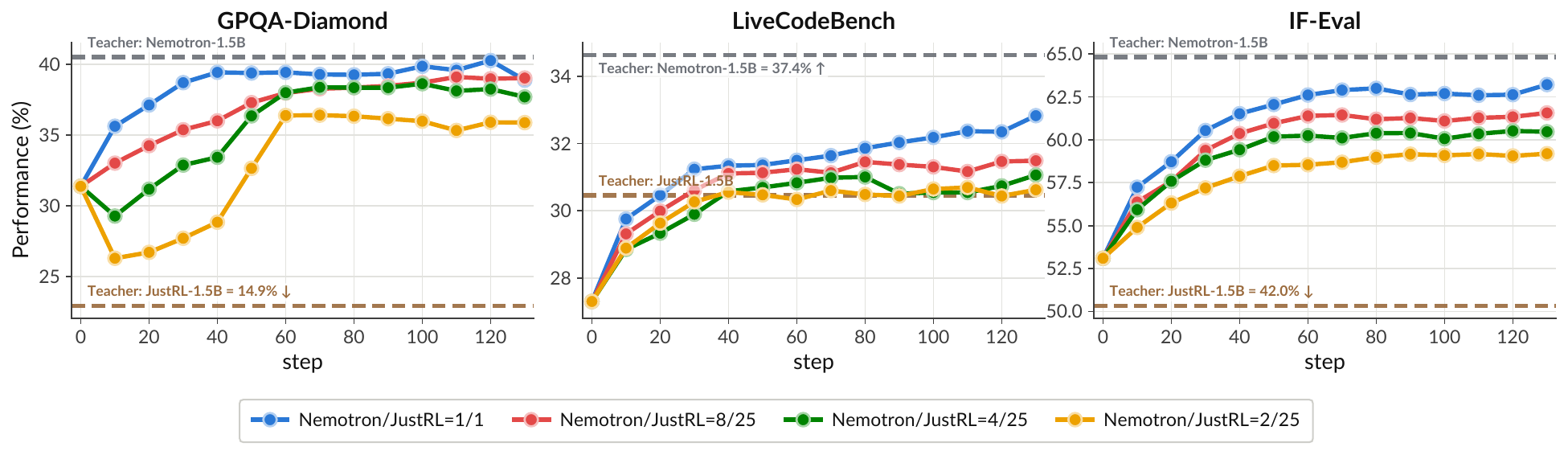}
        \caption{Setting 1. Math teacher: JustRL-1.5B; Science/IF teacher: Nemotron-1.5B; student: Dev-1.5B.}
        \label{fig:mopd_nemo10step}
    \end{subfigure}

    \begin{subfigure}[b]{0.99\linewidth}
        \centering
        \includegraphics[width=\linewidth]{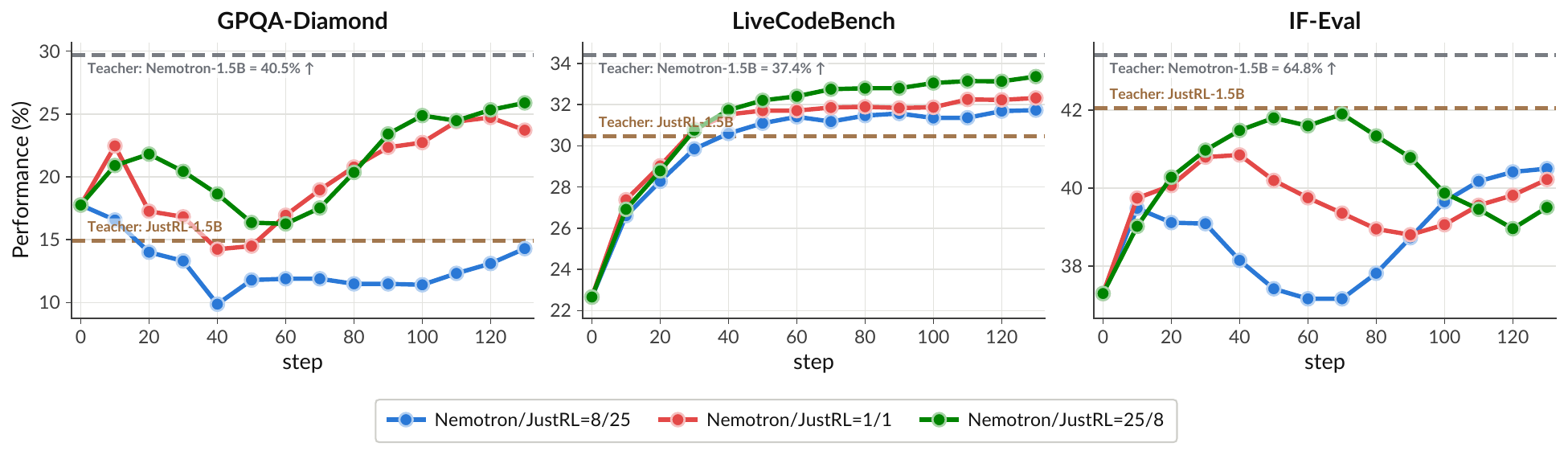}
        \caption{Setting 2. Math teacher: Nemotron-1.5B; Science/IF teacher: JustRL-1.5B; student: DS-distill-1.5B.}
        \label{fig:mopd_reverse_teacher}
    \end{subfigure}

    \caption{\textbf{\emph{The MOPD seesaw effect}}. Changing the mixture ratio of different teachers' allocated prompts significantly changes the OPD students' capabilities in different domains. For example, in subfigure (a), student's GPQA-Diamond (scientific reasoning), LiveCodeBench (code reasoning), and IF-Eval (instruction-following) performance gradually drops down as we increase the proportion of JustRL-1.5B (math teacher) data.}
    \label{fig:mopd_seesaw_effect}
\end{figure*}
\subsection{Cross-Domain Interference in MOPD}
\label{subsec:mopd}

\begin{table}[t]
\centering
\caption{MOPD math accuracy for different JustRL-1.5B/Nemotron-1.5B mixture ratios (J/N).}
\label{tab:mopd-step200}
\resizebox{0.6\columnwidth}{!}{%
\begin{tabular}{c|cc|c}
\toprule
\textbf{Config} & \textbf{BeyondAIME} & \textbf{OlymMATH} & \textbf{Average} \\
\midrule
\midrule
DS-distill-1.5B & $9.6\%$ & $14.2\%$ & $11.9\%${\footnotesize \textcolor{black!60!black}{(\textbf{student})}} \\
\midrule
\midrule
\multicolumn{4}{c}{\textbf{\textit{Math Teacher: JustRL-1.5B, Science/IF Teacher: Nemotron-1.5B}}} \\
\midrule
$\mathbf{\text{J}/\text{N}=25/2}$ & $19.4\%$ & $33.3\%$ & $26.4\%${\footnotesize \textcolor{green!60!black}{(\textbf{+0.1 pp})}} \\
$\mathbf{\text{J}/\text{N}=25/8}$ & $19.6\%$ & $34.5\%$ & $27.1\%${\footnotesize \textcolor{green!60!black}{(\textbf{+0.8 pp})}} \\
$\mathbf{\text{J}/\text{N}=1/1}$ & $20.2\%$ & $32.4\%$ & $26.3\%${\footnotesize \textcolor{black!60!black}{(\textbf{baseline})}} \\
$\mathbf{\text{J}/\text{N}=8/25}$ & $19.7\%$ & $32.5\%$ & $26.1\%${\footnotesize \textcolor{red!60!black}{(\textbf{-0.2 pp})}} \\
$\mathbf{\text{J}/\text{N}=2/25}$ & $19.2\%$ & $31.0\%$ & $25.1\%${\footnotesize \textcolor{red!60!black}{(\textbf{-1.2 pp})}} \\
\midrule
\midrule
\multicolumn{4}{c}{\textbf{\textit{Math Teacher: Nemotron-1.5B, Science/IF Teacher: JustRL-1.5B}}} \\
\midrule
$\mathbf{\text{J}/\text{N}=25/8}$ & $17.4\%$ & $28.0\%$ & $22.7\%${\footnotesize \textcolor{green!60!black}{(\textbf{+0.7 pp})}} \\
$\mathbf{\text{J}/\text{N}=1/1}$ & $17.3\%$ & $26.6\%$ & $22.0\%${\footnotesize \textcolor{black!60!black}{(\textbf{baseline})}} \\
$\mathbf{\text{J}/\text{N}=8/25}$  & $17.1\%$ & $26.2\%$ & $21.7\%${\footnotesize \textcolor{red!60!black}{(\textbf{-0.3 pp})}} \\
\bottomrule
\end{tabular}%
}
\end{table}

\begin{figure*}[t]
    \centering
    \captionsetup[subfigure]{font=scriptsize}
    \begin{subfigure}[b]{0.32\linewidth}
        \centering
        \includegraphics[width=\linewidth]{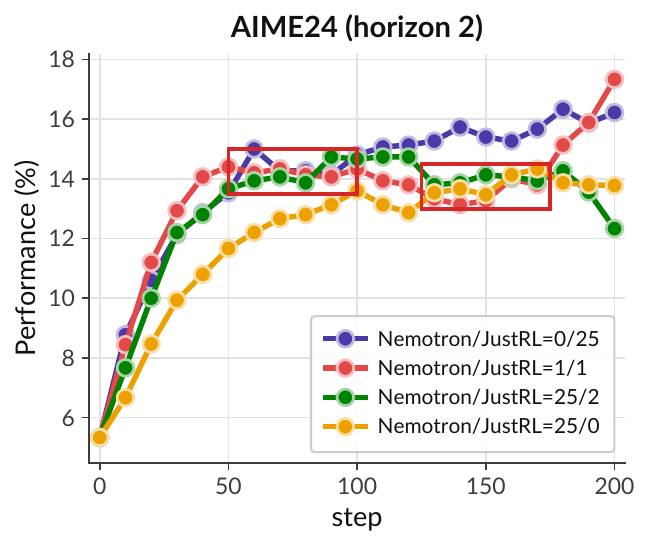}
        \caption{JustRL-Math, Nemotron-Science/IF}
        \label{fig:switch_mopd}
    \end{subfigure}
    \hfill
    \begin{subfigure}[b]{0.32\linewidth}
        \centering
        \includegraphics[width=\linewidth]{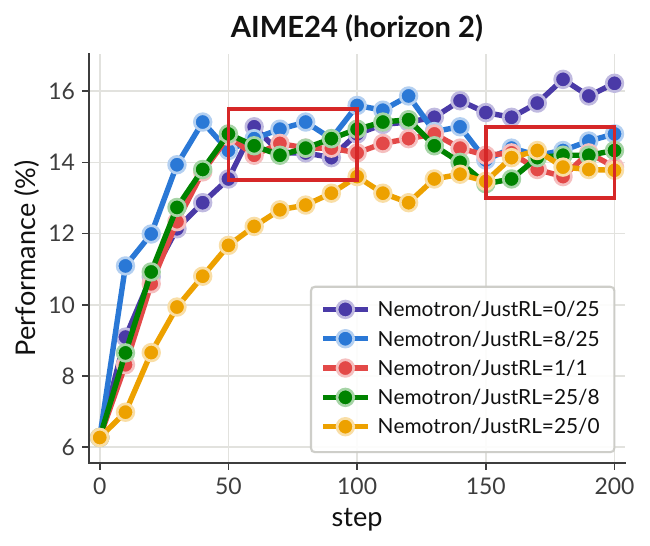}
        \caption{JustRL-Science/IF, Nemotron-Math}
        \label{fig:switch_mopd_reverse_teacher}
    \end{subfigure}
    \hfill
    \begin{subfigure}[b]{0.32\linewidth}
        \centering
        \includegraphics[width=\linewidth]{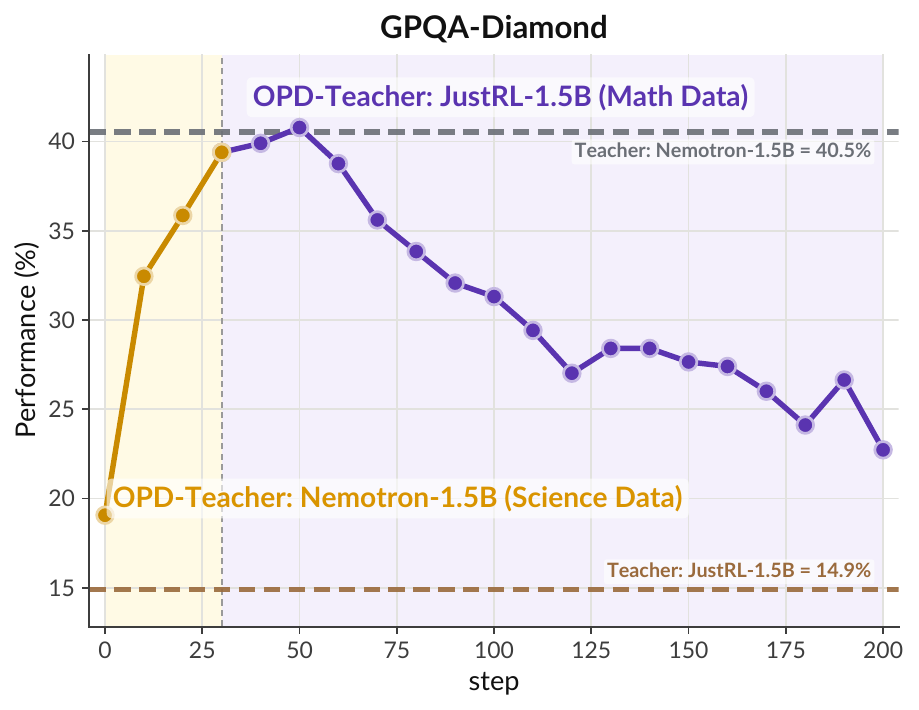}
        \caption{Cascaded OPD, GPQA-Diamond}
        \label{fig:cascaded_opd}
    \end{subfigure}
    \caption{A tug-of-war between the teachers can be observed during MOPD training. (a,b) \textcolor{red}{highlighted red boxes} show that the MOPD student's performance \textbf{\emph{first tracks}} the \textcolor{violet}{JustRL-only training curve} and \textbf{\emph{later drifts toward}} the \textcolor{yellow!85!black}{Nemotron-only training curve}, under both settings. (c) In cascaded OPD (one teacher then the other), GPQA performance \textcolor{yellow!85!black}{rises sharply} under the guidance of Nemotron-1.5B with science prompts and then \textcolor{violet}{drops back} by the subsequent JustRL-1.5B with math prompts, decoupling the effect of multi-teacher in the temporal dimension.}
    \label{fig:additional_result_mopd}
\end{figure*}

The single-teacher cross-domain generalization observations have a critical but easily overlooked implication for Multi-teacher OPD (MOPD): because a teacher affects evaluation domains far beyond that of its OPD training prompts, \textbf{\emph{routing each prompt to a domain teacher does not keep that teacher's influence within its assigned domain}}. This raises a practical question for MOPD: whether teachers assigned to different domains \emph{\textbf{contend for the same capability}}, as each teacher's cross-domain influence may overlap with, and be pulled against, another teacher's supervision in its assigned domain.
We test this with two students, Dev-1.5B\footnote{See Appendix~\ref{sec:appendix_model_lineage} for detailed introduction} and DS-distill-1.5B and two teachers with complementary profiles, JustRL-1.5B (stronger on math) and Nemotron-1.5B (much stronger on science and IF, weaker on math). In Setting~1, JustRL-1.5B is the math teacher and Nemotron-1.5B is the science/IF teacher. To isolate the effect of the routed prompt domain and to further probe the seesaw effect, Setting~2 swaps their roles, using Nemotron-1.5B as the math teacher and JustRL-1.5B as the science/IF teacher, and we track how the student's per-domain performance changes.

\paragraph{Teacher Mixture Ratios Induce a Seesaw Effect.}
As we change the ratio of prompts allocated to the two teachers while keeping their total amount unchanged, the student's performance on each benchmark moves toward the teacher that receives the larger share rather than being decided by which domain the teacher is assigned to teach. 
In Setting~1 (Figure~\ref{fig:mopd_nemo10step}), as we increase JustRL's (the math teacher) share (Nemotron/JustRL changes from $1/1$ to $2/25$), the student's GPQA-Diamond, LiveCodeBench, and IFEval scores all decline toward JustRL's lower scores; \textbf{\emph{on GPQA-Diamond the pull from the math teacher JustRL is strong enough that accuracy even drops $\sim5\%$ early in training}}, because JustRL scores below the student Dev-1.5B there. In Setting~2 (Figure~\ref{fig:mopd_reverse_teacher}), where Nemotron is instead assigned as the math teacher, the same three benchmarks now \emph{rise} as we raise Nemotron's share, moving toward Nemotron's higher scores on them. 
The math results (Table~\ref{tab:mopd-step200}) give the complementary case, where properly increasing the JustRL share (JustRL's math performance is better than Nemotron) improves math regardless of its assigned domain.

Ultimately, the performance shift is not merely dictated by the teacher's assigned domain: \textbf{\emph{increasing a teacher's share uniformly pulls multi-domain capabilities toward that teacher's baseline, creating a mixture-dependent seesaw that renders domain-specific prompt routing ineffective at confining a teacher's influence}}.


\paragraph{A tug-of-war between the teachers can be observed during training.}
Finally, the balance between teachers shifts over training. On AIME24-Horizon-2 under both two settings (Figures~\ref{fig:switch_mopd} and~\ref{fig:switch_mopd_reverse_teacher}), MOPD students first track the stronger-math JustRL-only curve and later drift toward the lower Nemotron-only level, so the student does not settle immediately at a fixed mixture outcome but is pulled between the two teachers as training proceeds. 
In Figure~\ref{fig:switch_mopd}, where JustRL is the math teacher, we can observe that finally the red curve (Nemotron/JustRL mixture ratio $1/1$) is re-pulled back to the JustRL-only curve, while the yellow curve (Nemotron/JustRL mixture ratio $25/2$, fewer prompts allocated to JustRL) is entirely pulled towards the Nemotron-only curve.
We additionally conduct a cascaded OPD experiment (Figure~\ref{fig:cascaded_opd}) to decouple the effect of multi-teacher OPD training in the temporal dimension: training first with the science-strong Nemotron-1.5B teacher and science prompts raises GPQA toward its level, and after switching to JustRL-1.5B and math prompts, GPQA moves back down toward the original level.


\paragraph{The Seesaw Effect Provides a Mental Model for Analyzing MOPD.}
The seesaw effect offers a useful perspective for analyzing the counteraction in MOPD~\citep{chen2026counteraction}: when the student underperforms on a domain, the cause need not lie with the teacher assigned to that domain, since another teacher can pull the same capability through its own cross-domain transfer. This suggests examining the full set of teachers, rather than only the domain expert in question, when diagnosing MOPD, and it also indicates that a domain expert's capabilities outside its target domain are relevant to how it behaves in MOPD, not only its performance on the assigned domain. The traditional prompt-routing paradigm~\citep{ma2026mopd,coreteam2026mimov2flashtechnicalreport,huang2026kat,xu2026deepseek} therefore does not ensure that different domain experts' capabilities stay isolated from one another.

\begin{insightbox}
\textbf{Takeaway:}
Because a teacher's influence is not confined to the domain of its OPD prompts, routing prompts to domain experts in multi-teacher OPD (MOPD) cannot isolate them. As the mixture ratio changes, the student's score on each benchmark moves toward the teacher with the larger share, and toward that teacher's own level rather than the domain it is assigned to teach. 
Combining experts therefore produces a mixture-dependent \emph{seesaw} among their capabilities rather than a clean sum of expert skills. This offers a useful perspective for diagnosing MOPD: underperformance on a domain may stem from another teacher's cross-domain pull, so a domain expert's off-domain behavior also matters.
\end{insightbox}

%% file: sections/discussion.tex
\section{More Discussion on Same/Cross-Origin OPD and MOPD Experiments}

\begin{figure*}[t]
    \centering
    \begin{subfigure}[b]{0.48\linewidth}
        \centering
        \includegraphics[width=\linewidth]{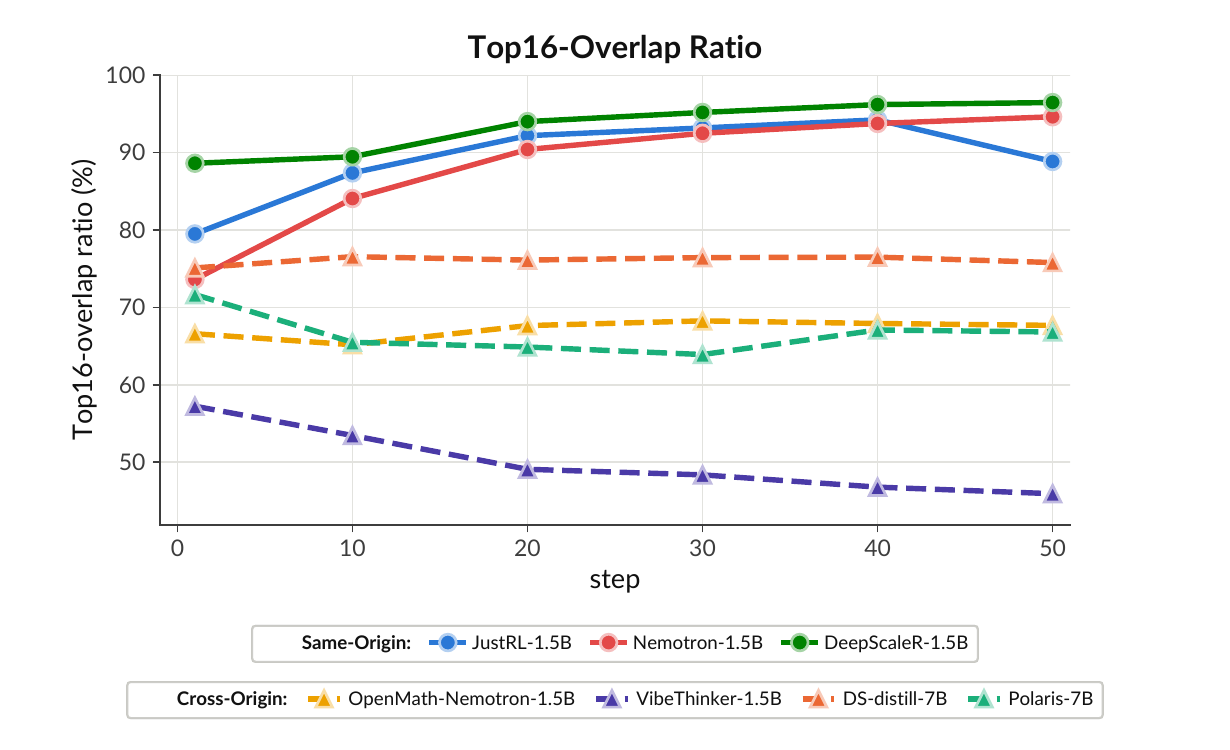}
        \caption{DS-distill-1.5B OPD Experiments}
    \end{subfigure}
    \begin{subfigure}[b]{0.48\linewidth}
        \centering
        \includegraphics[width=\linewidth]{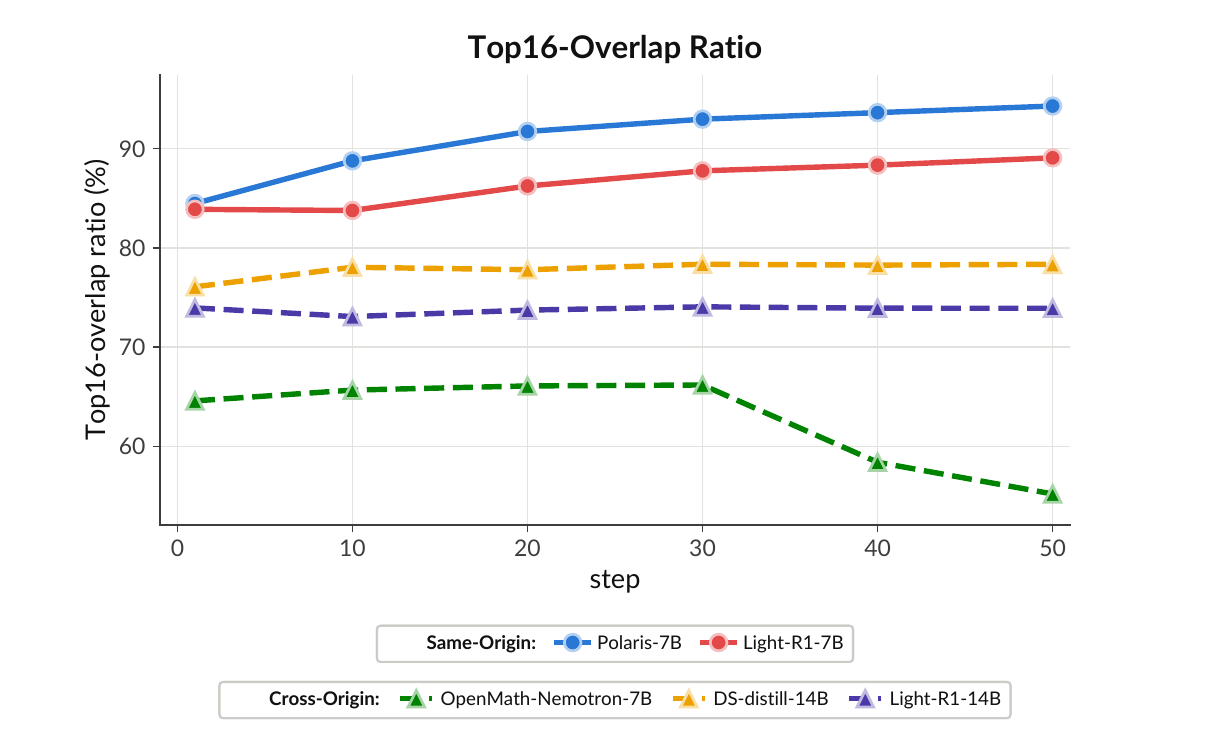}
        \caption{DS-distill-7B OPD Experiments}
    \end{subfigure}
    
    \caption{Top-K (K=16) Overlap Ratio of the teacher and student models in different OPD experiments. In Figure (a) and (b), the student models are DS-distill-1.5B and DS-distill-7B, respectively. The solid lines refer to OPD experiments with same-origin teachers; the dashed lines refer to OPD experiments with different-origin teachers.}
    \label{fig:top16_overlap_ratio}
\end{figure*}

\begin{figure}[htbp]
    \centering
    \includegraphics[width=0.99\linewidth]{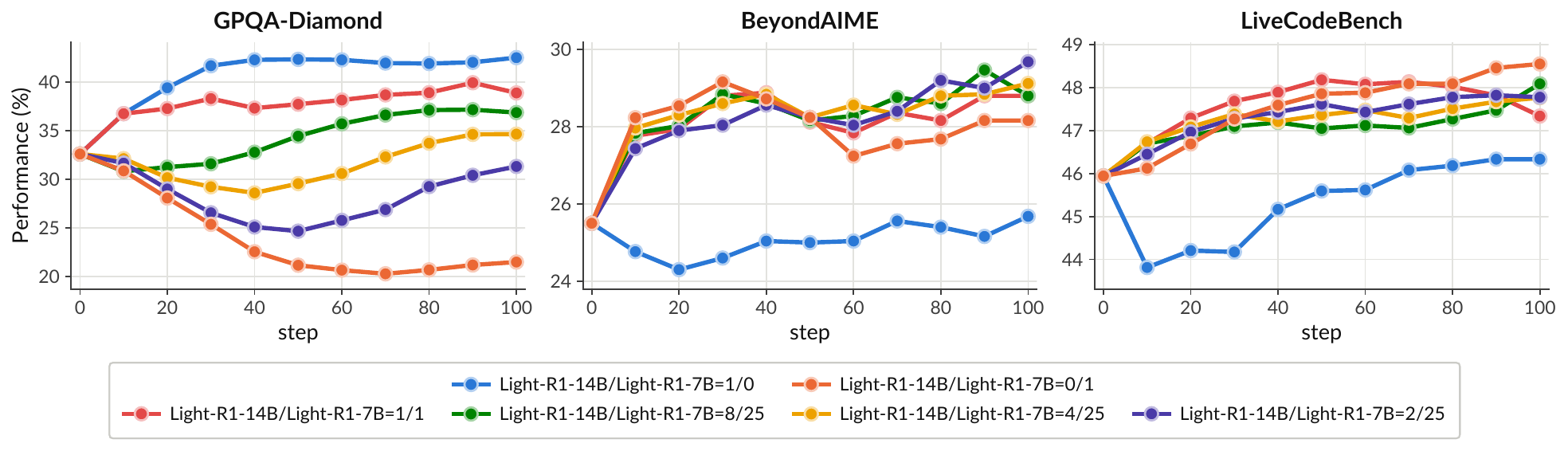}
    \caption{The MOPD experiment results of DS-distill-7B (student), Light-R1-7B (math teacher), and Light-R1-14B (science/IF teacher). The legends refer to the expert data mix ratio of different MOPD experiments (Light-R1-14B/Light-R1-7B $\in\{1/0, 1/1, 8/25, 4/25, 2/25, 0/1\}$).}
    \label{fig:mopd_7b}
\end{figure}

Throughout the paper, model origin recurs as the factor that most consistently separates broad generalization from narrow fitting. We close with two analyses that look more directly at why: how OPD reshapes the student's policy over training, and how this plays out when a same-origin and a cross-origin teacher compete within MOPD.

\paragraph{Same-origin OPD aligns the student's policy as a whole.}
To probe how OPD changes the student's policy beyond the training loss, we measure the top-$K$ ($K{=}16$) overlap ratio~\citep{li2026rethinking} between the teacher's and the student's next-token distributions during training (Figure~\ref{fig:top16_overlap_ratio}).
Two patterns are consistent across the 1.5B and 7B students.
First, the overlap ratio starts clearly higher for same-origin teachers than for cross-origin ones, confirming that a shared origin already places the student's policy closer to the teacher's before any OPD.
Second, and more telling, the same-origin overlap ratio rises markedly over training, whereas the cross-origin overlap ratio stays flat or even declines.
Since both settings minimize the same teacher--student KL objective, this contrast indicates that same-origin OPD progressively aligns the student to the teacher's policy \emph{as a whole}, while cross-origin OPD reduces the divergence on the training distribution without pulling the two policies into broader agreement.
This offers a mechanistic reading of ``same-origin generalizes, cross-origin fits'': broad transfer follows from whole-policy alignment, which is far easier to achieve when teacher and student share an origin.

\paragraph{A same-origin teacher exerts stronger pull in MOPD.}
The same asymmetry surfaces when a same-origin and a cross-origin teacher are combined. We repeat the MOPD experiment on DS-distill-7B, using the same-origin Light-R1-7B as the math teacher and the cross-origin Light-R1-14B as the science/IF teacher (Figure~\ref{fig:mopd_7b}).
On GPQA-Diamond, the student's science accuracy is readily pulled toward the math teacher Light-R1-7B, even at a balanced $1{:}1$ mixture where the science/IF teacher receives an equal share.
Read together with the alignment analysis above, this suggests that a same-origin teacher exerts a stronger pull on the student than a cross-origin one in MOPD, so that the seesaw between teachers is tilted not only by the mixture ratio but also by how close each teacher's origin is to the student.


\begin{figure}[htbp]
    \centering
    \includegraphics[width=0.99\linewidth]{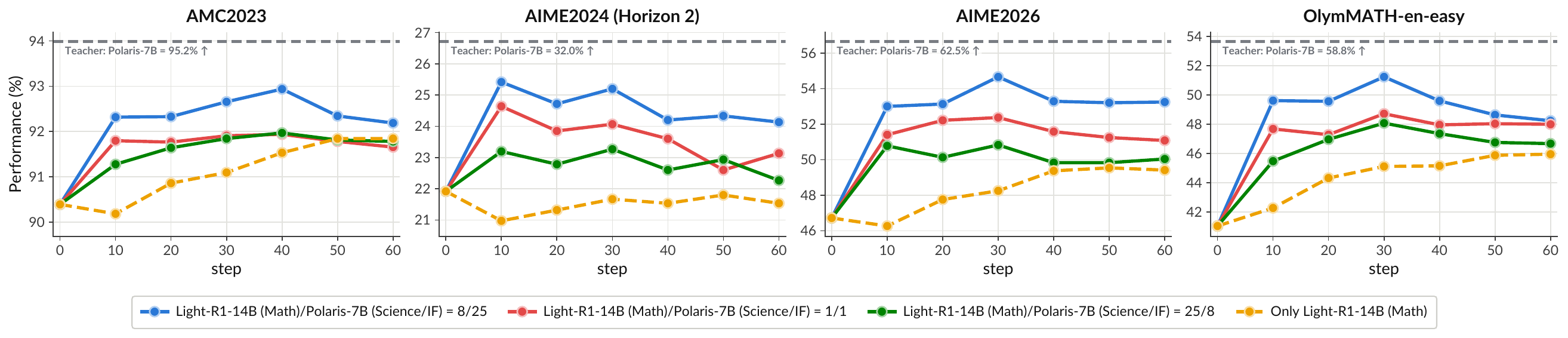}
    \caption{MOPD on DS-distill-7B (student) with Light-R1-14B as the math expert and Polaris-7B as the science/IF expert, evaluated on four math benchmarks (AMC2023, AIME2024 (Horizon 2), AIME2026, OlymMATH-en-easy). Solid lines are MOPD runs at math/science-IF data mix ratios Light-R1-14B/Polaris-7B $\in\{8/25, 1/1, 25/8\}$; the dashed orange line is single-teacher OPD with only the math expert Light-R1-14B; the dashed grey line marks the science/IF expert Polaris-7B's own math accuracy.}
    \label{fig:mopd_polaris_light}
\end{figure}

\paragraph{A MOPD instance shows that cross-domain transfer, not the math data share, drives the math performance gains of the student model.}
The pull of origin can even invert the effect of the mixture ratio. We run MOPD on DS-distill-7B with the cross-origin Light-R1-14B as the math expert and the same-origin Polaris-7B as the science/IF expert, and evaluate math ability across four benchmarks (Figure~\ref{fig:mopd_polaris_light}).
Counterintuitively, \emph{raising} the math data share does not help math: as the Light-R1-14B/Polaris-7B mixture moves from $8/25$ toward $25/8$, math accuracy on all four benchmarks steadily \emph{drops} rather than rises, so pouring in more of the math expert's data yields worse math---the opposite of what a data-centric view would predict.
Two references make the mechanism clear. First, the science/IF expert Polaris-7B, despite carrying no math data, is itself a very strong math model (grey line; e.g.\ $95.2\%$ on AMC2023 and $62.5\%$ on AIME2026), so its share of the mixture transfers math ability to the same-origin student for free.
Second, single-teacher OPD from the math expert Light-R1-14B alone (dashed orange) is the \emph{weakest} of all runs on every benchmark, confirming that the cross-origin math teacher transfers poorly on its own.
The math gains therefore come not from the math expert's data but from the same-origin science/IF expert, whose whole-policy alignment carries math along with it; adding more of the cross-origin math data merely displaces this effective same-origin signal. This is the same origin-driven pull as above, now strong enough to reverse the sign of the mixture-ratio effect, and underscores how central cross-origin transfer is to what MOPD actually learns.

\begin{figure*}[t]
    \centering
    \begin{subfigure}[b]{0.32\linewidth}
        \centering
        \includegraphics[width=\linewidth]{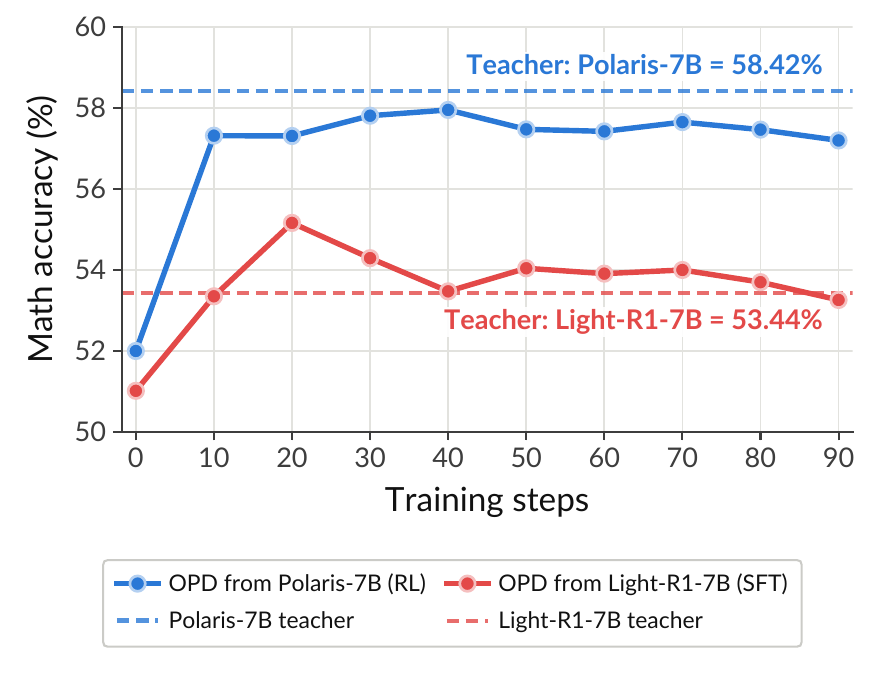}
        \caption{English math performance change.}
    \end{subfigure}
    \begin{subfigure}[b]{0.64\linewidth}
        \centering
        \includegraphics[width=\linewidth]{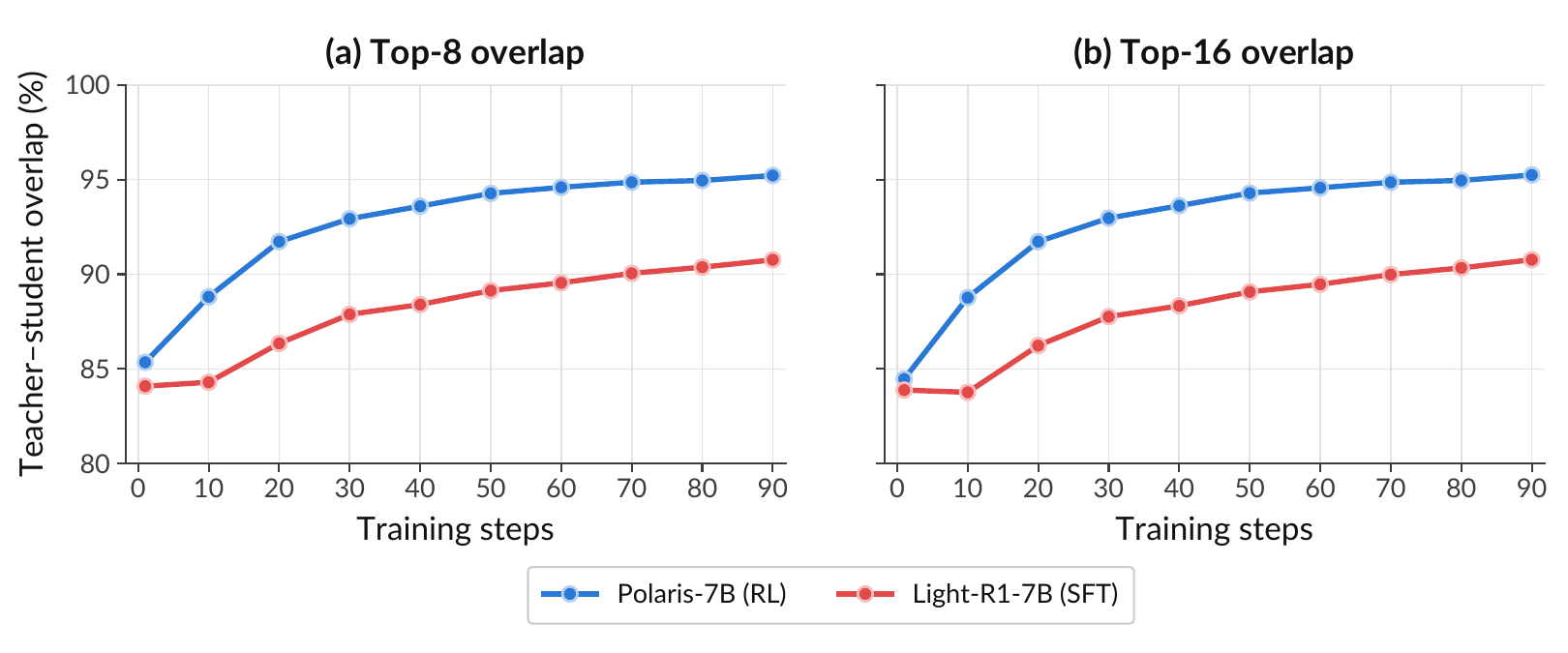}
        \caption{Top-k overlap ratio change.}
    \end{subfigure}
    
    \caption{A preliminary comparison between RL-teacher and SFT-teacher (same-origin setting).}
    \label{fig:sft_vs_rl}
\end{figure*}
\paragraph{A preliminary comparison between the OPD effect of the RL-teacher and the SFT-teacher.} SFT, unlike RL, is an off-policy training paradigm, which can reshape the model's representation more drastically~\citep{chen2026retainingdoingroleonpolicy}. To this end, we also super wonder whether using SFT-teacher in the same-origin OPD training leads to totally different results or not.
Again, we use the DS-distill-7B as the student model, use the Polaris-7B as the RL-teacher, and use the Light-R1-7B as the SFT-teacher.
The results (average performance over the six main English math reasoning benchmarks and top-8/16 teacher-student overlap ratio) are shown in Figure~\ref{fig:sft_vs_rl}.
Our key observations include: (1) in terms of the top-k overlap ratio, although the ratios both increase remarkably, RL-teacher is much better at align the student and teacher; (2) in terms of the benchmark performance, both SFT and RL same-origin teachers can make student achieve the teacher-like performance after the OPD-training.
So our preliminary conclusion is that SFT/RL same-origin teachers are not playing different roles in the OPD in our experiment setting.
However, the readers should notice that \textbf{\textit{SFT teachers' OPD effect also largely depends on the distribution of the introduced off-policy SFT data}} (e.g., in-distribution with the students' seen training data or with huge distribution-shift?) and \textit{\textbf{the extent of the SFT training}} (e.g., only a few steps or several epochs of huge data training?). 
In the real large-scale training scenarios, we often observe that OPD with SFT teachers can not fully recover the teacher's expert performance.
\begin{insightbox}
\textbf{Takeaway:} 
Same-origin OPD raises the teacher--student top-$K$ overlap over training, whereas cross-origin OPD does not, so broad generalization arises from aligning the student's policy \emph{as a whole} rather than merely lowering the KL on the training distribution. Consequently, a same-origin teacher pulls the student more strongly than a cross-origin one, tilting the MOPD seesaw by origin as well as by mixture ratio: so that adding more cross-origin math data can \emph{lower} math accuracy, while a same-origin non-math expert transfers math ability freely.
\end{insightbox}

%% file: appendix/appendix.tex
\appendix
\section{Related Work}
\label{appendix:related_work}

\subsection{OPD in Large-Scale Post-Training}


Conventional knowledge distillation typically trains a student on
teacher-generated responses, creating a mismatch between the sequences
observed during training and those generated by the student at inference
time.
GKD reduces this mismatch by obtaining teacher feedback on
student-generated sequences~\cite{agarwal2024policy}.
MiniLLM further formulates language-model distillation with reverse KL
optimization and develops an on-policy policy-gradient estimator
\cite{gu2024minillm}.
More recently, Thinking Machines Lab presented OPD as a practical
post-training paradigm that combines student-side exploration with dense
token-level teacher supervision~\cite{lu2025onpolicydistillation}.


OPD and its multi-teacher variants have subsequently become important
components of large-scale LLM post-training.
Recent technical reports employ them to consolidate independently
trained domain experts and integrate diverse abilities
into a single model~\cite{
coreteam2026mimov2flashtechnicalreport,
xu2026deepseek,
yang2026nemotron,
huang2026kat,
team2026mach}.
These studies provide strong evidence for the practical effectiveness
and scalability of OPD.
However, they primarily evaluate the final integrated models and do not
systematically isolate whether the distilled capabilities generalize~\citep{li2026rolereasoningpatternsgeneralization,ren2026rethinkinggeneralizationreasoningsft,chu2025sft,garcia2025exploringllmscapturerepresent}
under controlled changes in training and evaluation distributions.

\subsection{Understanding OPD mechanisms}


Recent studies have investigated the conditions under which OPD succeeds.
\citet{li2026rethinking} identifies compatible teacher--student thinking patterns
and genuinely novel teacher capabilities as two important factors for
effective distillation.


Other work focuses on the optimization pathologies of OPD.
\citet{fu2026revisiting} analyzes the bias and variance of sampled-token
optimization and identifies unreliable guidance on student-generated
prefixes and tokenizer mismatch as major failure modes.
A complementary study characterizes OPD as an exploration mechanism and
highlights student--teacher mismatch and length exploitation as two
central pathologies~\cite{wang2026demystifying}.


Several methods improve OPD by modifying its rollout or optimization
procedure.
Early-Stopping OPD limits supervision to earlier response positions,
where teacher guidance is less affected by student-prefix drift
\cite{ziheng2026less}.
Asymmetric OPD applies different optimization treatments to positive and
non-positive token advantages to balance exploitation and imitation
\cite{jia2026asymmetric}.
While these studies explain and mitigate local training failures, we
focus on a complementary question: whether OPD transfers across
difficulty, language, reasoning horizon, task domain, and model origin.

\subsection{Data Selection in OPD}


Data selection has been extensively studied in reinforcement learning for LLM reasoning~\cite{yu2026dapo,kong2026rethinking,li2025learnalign}.
A growing body of work improves OPD by selecting or reweighting its supervision signals.
SCOPE separates correct and incorrect student trajectories, emphasizing
teacher corrective confidence on incorrect rollouts and student
uncertainty on correct ones~\cite{zheng2026scope}.
FiRe-OPD first filters unreliable trajectories and then softly reweights
informative tokens within the retained trajectories
\cite{li2026filter}.
BRTS samples multiple teacher trajectories and selects supervision based
on teacher correctness and alignment with the current student
\cite{zhang2026policy}.
Uni-OPD jointly considers student-side exploration and teacher-side
supervision reliability through a dual-perspective recipe
\cite{hou2026uni}.


These approaches select supervision at different granularities, but do not directly determine whether teacher-unsolved queries should be filtered or student-mastered queries should be retained.
We isolate these two factors through controlled teacher-side filtering
and student-side dynamic sampling, and further compare their effects across different model-origin settings.

\subsection{Capability Interaction in MOPD}


Multi-Teacher On-Policy Distillation (MOPD) integrates multiple
domain-specialized teachers into a single student by routing
student-generated trajectories to the corresponding teacher for
token-level supervision~\cite{ma2026mopd}.
By training domain experts independently before integration, MOPD
reduces the direct coupling among heterogeneous reinforcement-learning
objectives and provides a scalable approach to capability integration.


Nevertheless, supervision from multiple teachers may not be mutually
compatible~\citep{gu2026coevolvingpolicydistillation}.
CaMOPD identifies recovery--preservation counteraction caused by
conflicting teacher gradients and weak-signal flattening caused by
uniformly combining samples with different correction demands
\cite{chen2026counteraction}.
It addresses these issues through decoupled optimization and
teacher--student-gap-based sample selection.


Our work studies capability interaction from a complementary
cross-domain generalization perspective.
Rather than treating each teacher as transferring only its nominal
expert skill, we examine how its broader capability profile transfers
to both primary and non-primary domains.
Our generalization analysis might also be useful for understanding when OPD improvements transfer beyond their training setting~\citep{sang2026crispcompressedreasoningiterative,xu2026grpoonpolicydistillationempirical}.

\section{Experiment Settings}
\label{sec:appendix_exp_setting}

\subsection{Dataset}
\label{sec:appendix_datasets}

\subsubsection{Training Datasets}
\label{sec:appendix_training_datasets}

\paragraph{Mathematical reasoning.}

We use Big-Math-RL-Verified~\cite{albalak2025big} as the primary
mathematical training corpus.
For the absolute-difficulty comparison, the available GSM8K~\cite{cobbe2021traininggsm8k} and
DeepMath-103K-Hardest~\cite{he2026deepmathk} pools each contain approximately 8K queries.
We therefore sample a matched 8K-query subset from
Big-Math-RL-Verified to ensure that the comparison is not confounded by
the number of unique training queries.

\paragraph{Code reasoning.}

We use DeepCoder-Preview-Dataset~\cite{deepcoder2025} for
code-domain OPD.
The dataset contains approximately 24K competitive-programming problems
paired with executable test cases.
Its training set includes LiveCodeBench~\cite{jain2025livecodebench} problems submitted between
May~1, 2023 and July~31, 2024, together with verified problems from
TACO~\cite{li2023taco} and PrimeIntellect’s SYNTHETIC-1~\cite{2025synthetic1}.

\paragraph{Scientific reasoning.}

For the single-teacher cross-domain experiments, we use
TextbookReasoning~\cite{fan2025megascience} as the Science training corpus.
We remove all samples labeled as Mathematics or Computer Science to
reduce direct overlap with the Math and Code domains.
For the MOPD experiments, we follow the data domains used to train
Nemotron-Research-Reasoning-Qwen-1.5B~\cite{liu2026prorl}.
The science data are drawn from SCP-116K~\cite{lu2025scp}, after
filtering out all samples categorized as Mathematics.

\paragraph{Instruction following.}

The instruction-following data used in MOPD are taken from the
RL/instruction\_following split of
Llama-Nemotron-Post-Training-Dataset~\cite{bercovich2025llamanemotron}.
We use these examples together with the filtered SCP-116K data to
construct the science/IF training pool.
science and instruction-following examples each account for $50\%$
of the science/IF pool.

\subsubsection{Evaluation Benchmarks}
\label{sec:appendix_evaluation_datasets}

\paragraph{English mathematical reasoning.}

The English Math score is the average score over AMC~2023~\cite{MAA_AMC},
MATH-500~\cite{hendrycks2021measuringmath500}, AIME~2025/2026~\cite{MAA_AIME}, BeyondAIME~\cite{bytedance_seed_2025_beyondaime}, and OlymMATH-Hard~\cite{sun2026olymmath}.

\paragraph{Chinese mathematical reasoning.}

The Chinese Math score is the average score over OlymMATH-ZH~\cite{sun2026olymmath} and
LiveMathBench-ZH~\cite{liu2025livemathbench}.

\paragraph{Long-horizon mathematical reasoning.}

The Long-Horizon Math score is the average score over
AIME24-Horizon-2 and AMC23-Horizon-4 from R-HORIZON~\cite{lu2025r}.

\paragraph{Code reasoning.}

We evaluate code generation on the LiveCodeBench v5 problems published
between August~1, 2024 and February~1, 2025~\cite{jain2025livecodebench}, following the evaluation adopted by DeepCoder.

\paragraph{Science and instruction following.}

We evaluate scientific reasoning on GPQA-Diamond~\cite{rein2023gpqa} and instruction
following on IFEval~\cite{zhou2023instructionfollowingevaluationlargelanguage}.
GPQA-Diamond contains expert-level questions in physics, chemistry, and
biology, while IFEval evaluates compliance with programmatically
verifiable instructions.

\subsection{Model}
\label{sec:appendix_models}

\subsubsection{Model Lineage}
\label{sec:appendix_model_lineage}

We use model lineage to 
distinguish the difference between same-origin and cross-origin OPD.
Specifically, two models are considered same-origin when they
share the same concrete initialization checkpoint and one or both are
obtained by further post-training from that checkpoint.

Under this definition, DeepSeek-R1-Distill-Qwen-1.5B~\cite{guo2025deepseekr1} forms one lineage
root.
JustRL-DeepSeek-1.5B~\cite{he2025justrl} is obtained by applying Math-oriented
reinforcement-learning post-training to this checkpoint, while
Nemotron-Research-Reasoning-Qwen-1.5B~\cite{liu2026prorl} is obtained through prolonged
multi-domain reinforcement learning from the same initialization.
Dev-1.5B is also initialized from
DeepSeek-R1-Distill-Qwen-1.5B and further trained through 10 steps of
science/IF OPD.
Therefore, these four models belong to the same lineage.
DeepSeek-R1-Distill-Qwen-7B and Polaris-7B~\cite{Polaris2025} form a second lineage, since
Polaris-7B is obtained by further post-training the 7B distilled
checkpoint.
Likewise, DeepSeek-R1-Distill-Qwen-14B and Light-R1-14B~\cite{wen2025lightr1} form a third
lineage.
For the Qwen3 models, Qwen3-4B~\cite{yang2025qwen3} and Polaris-4B belong to the same
lineage because Polaris-4B is post-trained from Qwen3-4B.
Qwen3-8B and Qwen3-32B are treated as separate lineage roots because
neither is obtained by post-training the other.

\begin{table*}[t]
    \centering
    \small
    \caption{
    Post-training lineages of the models used in our experiments.
    Models sharing the same lineage root are treated as same-origin.
    }
    \label{tab:model_lineage}
    \resizebox{\linewidth}{!}{
    \begin{tabular}{llll}
        \toprule
        \textbf{Model} &
        \textbf{Lineage Root} &
        \textbf{Post-Training from the Root} &
        \textbf{Experimental Role} \\
        \midrule

        DS-R1-Distill-Qwen-1.5B
        & Qwen2.5-Math-1.5B
        & SFT on DeepSeek-R1 traces
        & Student \\

        JustRL-DeepSeek-1.5B
        & DS-R1-Distill-Qwen-1.5B
        & Math-oriented RL
        & Teacher \\

        DeepScaleR-1.5B
        & DS-R1-Distill-Qwen-1.5B
        & Math-oriented RL
        & Teacher \\
        
        Nemotron-Research-Reasoning-Qwen-1.5B
        & DS-R1-Distill-Qwen-1.5B
        & Multi-domain prolonged RL
        & Teacher \\

        Dev-1.5B
        & DS-R1-Distill-Qwen-1.5B
        & 10-step science/IF OPD
        & Student \\

        OpenMath-Nemotron-1.5B
        & Qwen2.5-Math-1.5B
        & SFT on OpenMathReasoning
        & Teacher \\

        VibeThinker-1.5B
        & Qwen2.5-Math-1.5B
        & SFT \& RLVR
        & Teacher \\
        
        \midrule
        
        DS-R1-Distill-Qwen-7B
        & Qwen2.5-Math-7B
        & SFT on DeepSeek-R1 traces
        & Student \\

        Polaris-7B
        & DS-R1-Distill-Qwen-7B
        & Math-oriented RL
        & Teacher \\

        Light-R1-7B
        & DS-R1-Distill-Qwen-7B
        & SFT
        & Teacher \\

        OpenMath-Nemotron-7B
        & Qwen2.5-Math-7B
        & SFT on OpenMathReasoning
        & Teacher \\
        
        \midrule
        
        DS-R1-Distill-Qwen-14B
        & Qwen2.5-14B
        & SFT on DeepSeek-R1 traces
        & Student \\

        Light-R1-14B
        & DS-R1-Distill-Qwen-14B
        & Long-CoT reasoning RL
        & Teacher \\

        \midrule

        Qwen3-4B
        & Qwen3-4B
        & None
        & Student \\

        Polaris-4B
        & Qwen3-4B
        & Math-oriented RL
        & Teacher \\

        \midrule

        Qwen3-8B-SFT
        & Qwen3-8B-Base
        & SFT on OpenThoughts3-1.2M
        & Student \\

        \midrule

        Qwen3-32B
        & Qwen3-32B
        & None
        & Teacher \\

        \bottomrule
    \end{tabular}
    }
\end{table*}

\subsection{Training Configuration}
\label{sec:training_config}

Most configurations converge within 100--200 steps, so we set
the maximum number of steps to 200.
The prompt batch size is 128, giving a maximum training budget of
$200\times128=25.6$K prompt instances per run.
For each prompt, the student generates four independent on-policy
responses, corresponding to a rollout group size of $N=4$.
The standard rollout decoding parameters are
temperature $1.0$, top-$p$ $1.0$, and unrestricted top-$k$
sampling, implemented as top-$k=-1$. We set the learning rate as $1e-5$ in all of the OPD/MOPD experiments.

The maximum sequence length depends on the teacher--student
configuration, as summarized in
Table~\ref{tab:max_sequence_length}.

\begin{table*}[t]
    \centering
    \caption{Maximum sequence lengths used for OPD rollout and evaluation.}
    \label{tab:max_sequence_length}
    \resizebox{0.8\linewidth}{!}{
    \begin{tabular}{lc}
        \toprule
        \textbf{Model Group} &
        \textbf{Maximum Length} \\
        \midrule

        Qwen3-4B, Qwen3-8B, Qwen3-32B, Polaris-4B
        & 40K \\

        DS-distill-1.5B,
        JustRL-1.5B, Nemotron-1.5B
        & 96K \\

        Light-R1-14B, DS-distill-14B
        & 64K \\

        Polaris-7B, DS-distill-7B 
        & 96K \\

        \bottomrule
    \end{tabular}
    }
\end{table*}

\subsection{Evaluation Configuration}

For benchmark evaluation, we use temperature $1.0$, top-$p$ $0.95$,
and top-$k=-1$.
The maximum sequence length follows
Table~\ref{tab:max_sequence_length}.
For each query, we independently sample $K$ responses and report
$\operatorname{Avg@}K$, defined as

\begingroup
\begin{align*}
    \operatorname{Avg@}K
    =
    \frac{1}{|\mathcal{D}|}
    \sum_{x\in\mathcal{D}}
    \frac{1}{K}
    \sum_{k=1}^{K}
    \mathcal{V}\!\left(x,y^{(k)}\right),
    \label{eq:avg_at_k}
\end{align*}
\endgroup

where $y^{(k)}$ is the $k$-th independently sampled response and
$\mathcal{V}$ is the benchmark-specific verifier.
Unlike Pass@$K$, Avg@$K$ evaluates every sampled response independently
and does not select the best response among the $K$ generations.

The number of evaluation samples used for each benchmark is summarized
in Table~\ref{tab:evaluation_sampling_num}.

\begin{table}[t]
    \centering
    \caption{
    Number of independently sampled responses used for benchmark
    evaluation.
    }
    \label{tab:evaluation_sampling_num}
    \resizebox{0.3\columnwidth}{!}{
    \begin{tabular}{lc}
        \toprule
        \textbf{Benchmark} & \textbf{Metric} \\
        \midrule

        AMC 2023
        & Avg@16 \\

        MATH-500
        & Avg@1 \\

        AIME 2025
        & Avg@16 \\

        AIME 2026
        & Avg@16 \\

        BeyondAIME
        & Avg@5 \\

        OlymMATH-Hard
        & Avg@5 \\

        OlymMATH-Easy
        & Avg@5 \\

        OlymMATH-ZH
        & Avg@5 \\

        R-HORIZON
        & Avg@10 \\

        LiveMathBench-ZH
        & Avg@16 \\

        GPQA-Diamond
        & Avg@4 \\

        LiveCodeBench v5
        & Avg@10 \\

        IFEval
        & Avg@4 \\

        \bottomrule
    \end{tabular}
    }
\end{table}

For visualization only, we smooth the evaluation curves using a
centered moving average.
Let $\{s_t\}$ denote the raw evaluation sequence and
$\widetilde{s}_t$ the displayed value.
We compute

\begingroup
\begin{align*}
    \widetilde{s}_t
    =
    \begin{cases}
        s_1,
        & t=1, \\[2pt]
        \displaystyle
        \frac{1}{3}\sum_{j=1}^{3}s_j,
        & t=2, \\[8pt]
        \displaystyle
        \frac{1}{5}\sum_{j=t-2}^{t+2}s_j,
        & t\geq 3.
    \end{cases}
\end{align*}
\endgroup

Thus, the first displayed point is left unsmoothed, the second point
averages the preceding, current, and following checkpoints, and all
subsequent points use a five-point centered moving average with radius
two.
The raw training trajectories extend beyond the final step shown in the
figures, so the complete five-point window is available for all displayed
points from the third point onward, including those near the right
boundary.
Centered moving averages are commonly used to suppress local
fluctuations while preserving the main trend of a sequence
\citep{hyndman2018forecasting}.
Smoothing is applied only for visualization and does not affect any
reported table value, checkpoint selection, or statistical calculation.

\subsubsection{Answer Extraction Instruction}
For all mathematical and scientific reasoning benchmarks, we add system prompts and extract the answer enclosed in the final \texttt{\textbackslash boxed\{\}} expression from each model response and use the Math-Verify\footnote{\url{https://github.com/huggingface/Math-Verify}} library to parse and compare the extracted answer with the reference answer.
For math tasks, we prompt the model to put the final answer (mostly the number and the mathematical expressions, sometimes the options) inside \texttt{\textbackslash boxed\{\}}.
\begin{insightbox}
\textbf{Math Reasoning Tasks (MATH-500, AMC23, AIME25/26, BeyondAIME, ...):}\\
Please reason step by step, and put your final answer within \texttt{\textbackslash boxed\{\}}.
\end{insightbox}
For GPQA-Diamond, we prompt the model to put the final choice (A, B, C or D) inside \texttt{\textbackslash boxed\{\}}.
\begin{insightbox}
\textbf{Scientific Reasoning Tasks (GPQA-Diamond):}\\
Please reason step by step, and put your final answer (only the option, i.e., A, B, C or D) within \texttt{\textbackslash boxed\{\}}.
\end{insightbox}
Note that these above two system prompts are consistent with all of the models used in this work.

\section{Additional Experiment Results}
\label{sec:appendix_add_results}
\subsection{Additional Results on the Training Data Difficulty}

To further test whether the observed behavior holds beyond difficulty partitions constructed from a single dataset, we introduce two additional datasets representing the extremes of the difficulty spectrum. For the extremely hard setting, we use DeepMath-103K~\cite{he2026deepmathk}, which provides difficulty scores ranging from 1 to 9, and retain queries with scores above 8. For the extremely easy setting, we use the GSM8K training set, which consists of grade-school mathematical word problems. We repeat the same OPD experiments on these two datasets to examine whether teacher-side pass rate becomes more important when the training queries are either almost always solvable or rarely solvable by the teacher.

\begin{figure*}[t]
    \centering
    \begin{subfigure}[b]{0.32\linewidth}
        \centering
        \includegraphics[width=\linewidth]{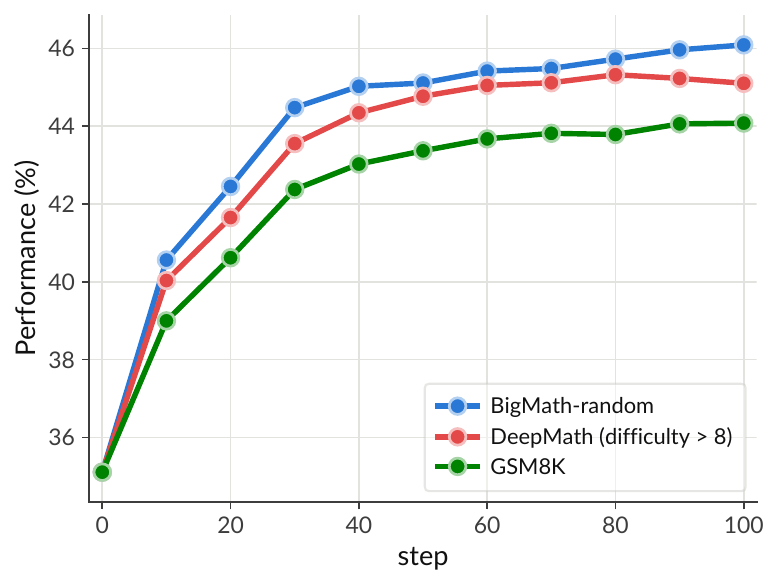}
        \caption{JustRL-1.5B $\rightarrow$ DS-distill-1.5B}
        \label{fig:difficulty_1d5b_to_1d5b}
    \end{subfigure}
    \hfill
    \begin{subfigure}[b]{0.32\linewidth}
        \centering
        \includegraphics[width=\linewidth]{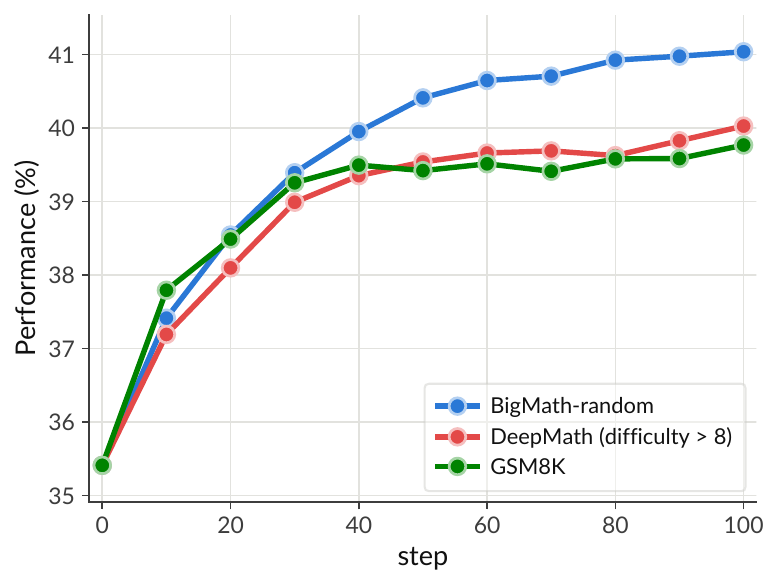}
        \caption{Polaris-7B $\rightarrow$ DS-distill-1.5B}
        \label{fig:difficulty_7b_to_1d5b}
    \end{subfigure}
    \hfill
    \begin{subfigure}[b]{0.32\linewidth}
        \centering
        \includegraphics[width=\linewidth]{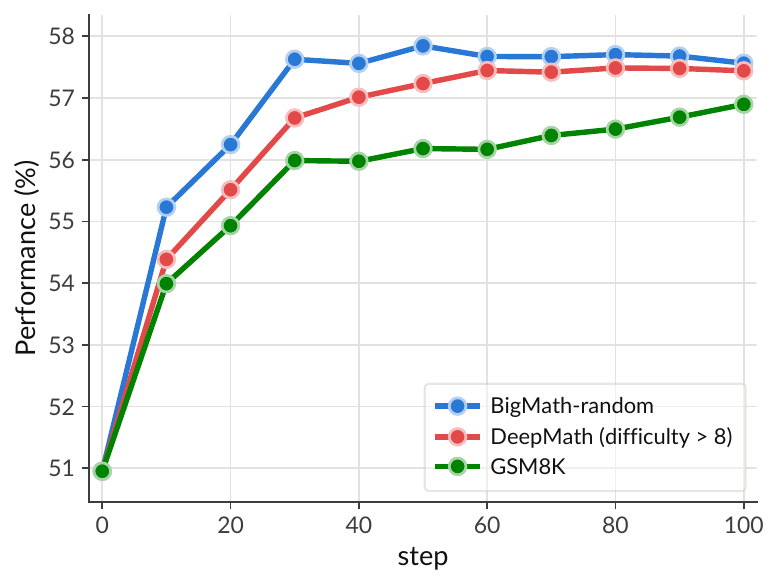}
        \caption{Polaris-7B $\rightarrow$ DS-distill-7B}
        \label{fig:difficulty_7b_to_7b}
    \end{subfigure}

    \caption{In-domain math performance (average over six English benchmarks) is largely insensitive to the difficulty of the training queries. Subfigure (a--c) training on the extremely easy GSM8K, the extremely hard DeepMath-103K, and the diverse BigMath mixture.}
    \label{fig:extreme_difficulty}
\end{figure*}

Figure~\ref{fig:extreme_difficulty} shows the reslts. The experiments on the two difficulty extremes lead to a similar conclusion.
Both grade-school-level GSM8K queries and highly challenging DeepMath-103K queries produce substantial improvements through OPD, and their final average scores differ by fewer than two points across the tested configurations.
Thus, effective on-policy supervision does not require the training queries to fall within a narrow absolute difficulty range.
However, models trained on either difficulty extreme generally remain behind those trained on the more diverse Big-Math-RL-Verified mixture.
This suggests that, although neither teacher pass rate nor absolute query difficulty alone determines whether a sample is useful, maintaining a diverse coverage of problem difficulties is still beneficial for broader generalization.

\subsection{Additional Generalization Results across Model Origin}
\label{sec:appendix_additional_generalization}

\begin{figure*}[t]
    \centering

    \begin{subfigure}[b]{0.99\linewidth}
        \centering
        \includegraphics[width=\linewidth]{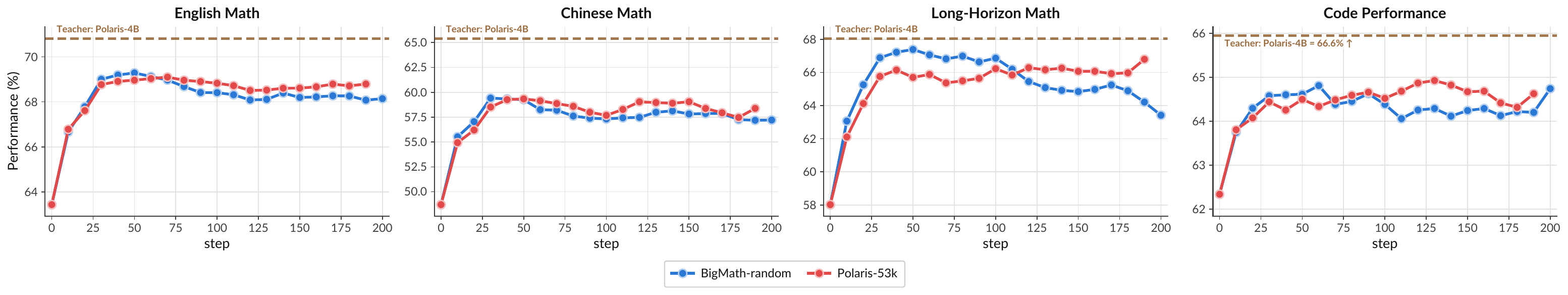}
        \caption{Polaris-4B to Qwen3-4B}
        \label{fig:4b_to_4b}
    \end{subfigure}
    
    \begin{subfigure}[b]{0.99\linewidth}
        \centering
        \includegraphics[width=\linewidth]{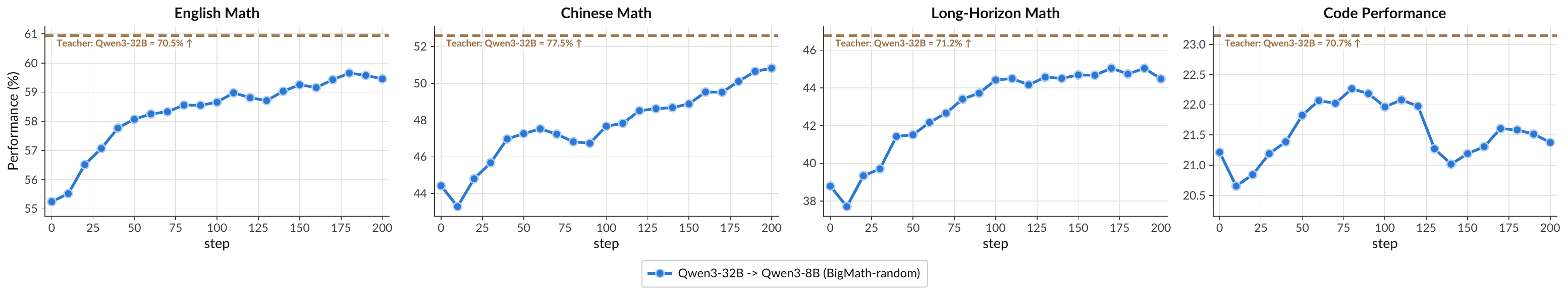}
        \caption{Qwen3-32B to Qwen3-8B-SFT}
        \label{fig:32b_to_8b}
    \end{subfigure}

     \begin{subfigure}[b]{0.99\linewidth}
        \centering
        \includegraphics[width=\linewidth]{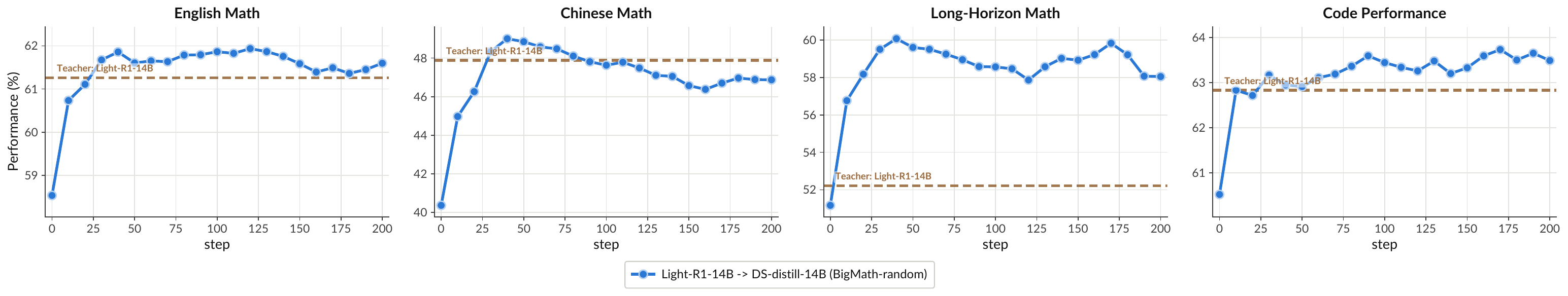}
        \caption{Light-R1-14B to DS-distill-14B}
        \label{fig:14b_to_14b}
    \end{subfigure}
    
    \caption{Additional generalization results across three teacher--student configurations.}
    \label{fig:math_generalization_and_code_generalization}
\end{figure*}

Figure~\ref{fig:math_generalization_and_code_generalization} presents
additional generalization results across three teacher--student
configurations.
Each panel reports performance on English Math, Chinese Math,
Long-Horizon Math, and Code, allowing us to jointly examine
in-domain distribution shifts and cross-domain transfer.

Figure~\ref{fig:4b_to_4b} shows the same-origin
Polaris-4B $\rightarrow$ Qwen3-4B configuration.
We compare OPD on the randomly sampled Big-Math-RL-Verified subset
with OPD on Polaris-53K, the dataset used for the post-training of
Polaris-4B.
Both training distributions produce stable improvements and converge
to similar performance across the evaluated benchmarks.
This result indicates that effective OPD does not require access to the teacher's original post-training data.
Figure~\ref{fig:32b_to_8b} reports the cross-origin
Qwen3-32B $\rightarrow$ Qwen3-8B-SFT configuration trained on
Big-Math-RL-Verified.
Qwen3-8B-SFT is initialized from Qwen3-8B-Base and supervised fine-tuned on OpenThoughts3-1.2M before OPD.
The student improves clearly on English Math, Chinese Math, and
Long-Horizon Math, demonstrating that the mathematical supervision
transfers across both language and reasoning-horizon shifts.
In contrast, the Code performance changes only marginally.
This result shows that under a cross-origin teacher--student configuration, broad in-domain generalization does not
necessarily imply equally strong transfer to every semantic domain.
Figure~\ref{fig:14b_to_14b} shows the same-origin
Light-R1-14B $\rightarrow$ DS-distill-14B configuration.
OPD produces stable improvements across all four evaluation groups.
Notably, on Long-Horizon Math, the distilled student eventually
surpasses the standalone teacher by a clear margin.
This observation suggests that the final student is not necessarily
bounded by the teacher's standalone benchmark accuracy.

\subsection{Complete Cross-domain Transfer Results}
\label{sec:appendix_complete_others_to_math}

\begin{figure*}[t]
    \centering
    \captionsetup[subfigure]{font=scriptsize}
    \begin{subfigure}[b]{0.32\linewidth}
        \centering
        \includegraphics[width=\linewidth]{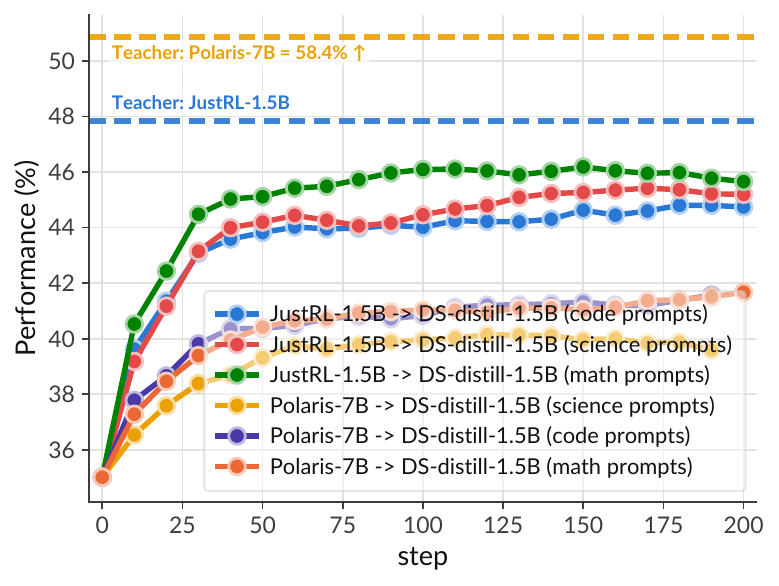}
        \caption{English Math (DS-distill-1.5B)}
        \label{fig:others_to_math_en_1d5b}
    \end{subfigure}
    \hfill
    \begin{subfigure}[b]{0.32\linewidth}
        \centering
        \includegraphics[width=\linewidth]{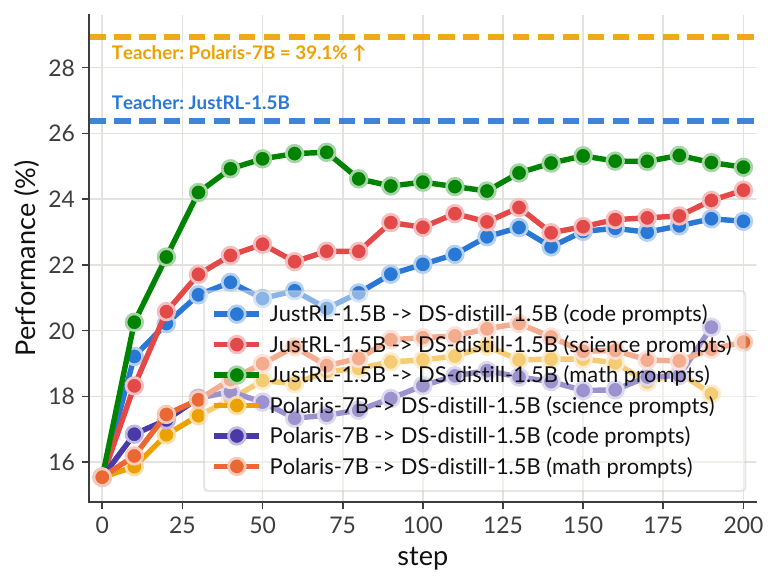}
        \caption{Chinese Math (DS-distill-1.5B)}
        \label{fig:others_to_math_zh_1d5b}
    \end{subfigure}
    \hfill
    \begin{subfigure}[b]{0.32\linewidth}
        \centering
        \includegraphics[width=\linewidth]{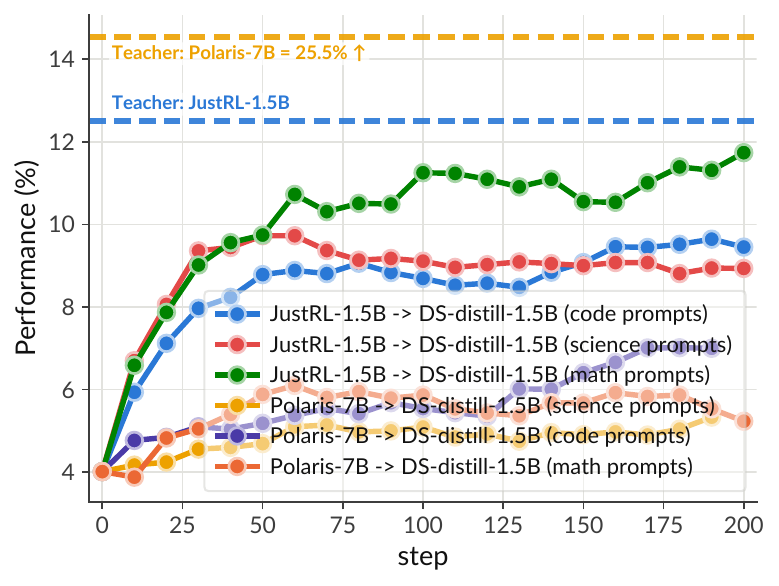}
        \caption{Long-Horizon Math (DS-distill-1.5B)}
        \label{fig:others_to_math_horizon_1d5b}
    \end{subfigure}

     \begin{subfigure}[b]{0.32\linewidth}
        \centering
        \includegraphics[width=\linewidth]{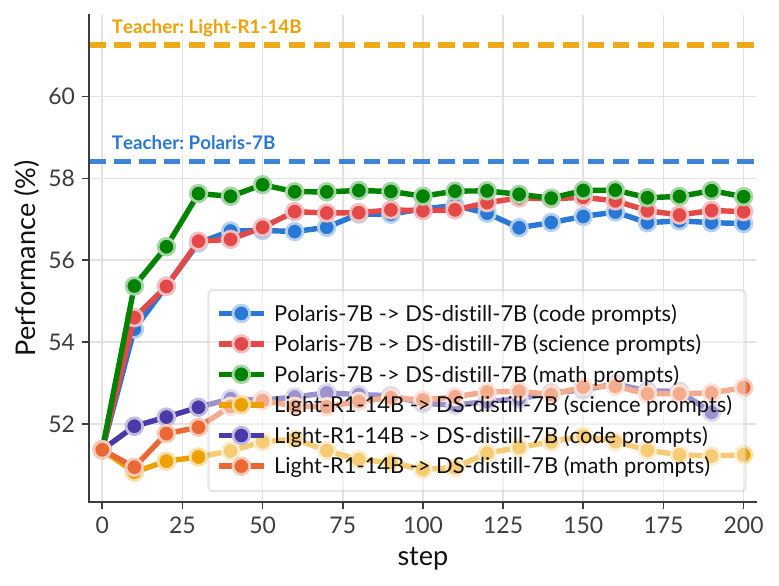}
        \caption{English Math (DS-distill-7B)}
        \label{fig:others_to_math_en_7b}
    \end{subfigure}
    \hfill
    \begin{subfigure}[b]{0.32\linewidth}
        \centering
        \includegraphics[width=\linewidth]{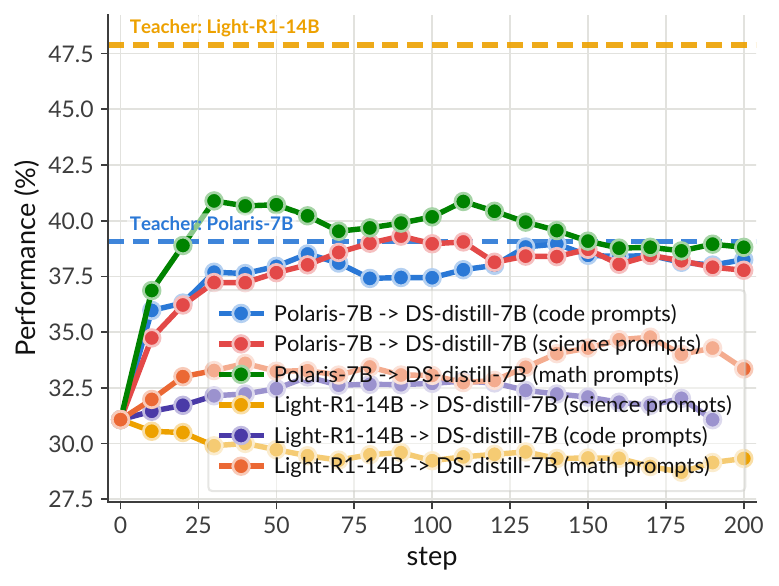}
        \caption{Chinese Math (DS-distill-7B)}
        \label{fig:others_to_math_zh_7b}
    \end{subfigure}
    \hfill
    \begin{subfigure}[b]{0.32\linewidth}
        \centering
        \includegraphics[width=\linewidth]{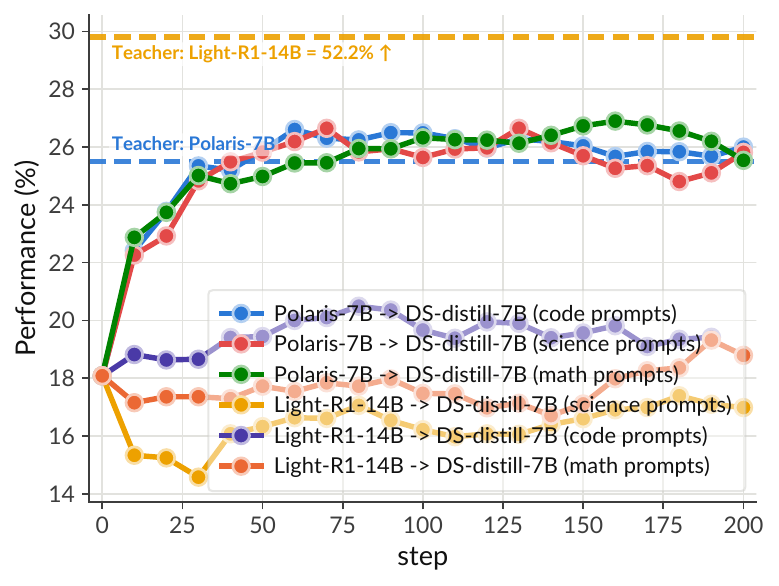}
        \caption{Long-Horizon Math (DS-distill-7B)}
        \label{fig:others_to_math_horizon_7b}
    \end{subfigure}
    
    \caption{Training on code and science domains and generalizing to math-related domains.}
    \label{fig:cross_domain_generalization_others_to_math}
\end{figure*}

Figure~\ref{fig:cross_domain_generalization_others_to_math} provides the
complete results for transferring from Code and Science training data to mathematical reasoning.
The first row presents results for DS-distill-1.5B, while the second row
presents the corresponding results for DS-distill-7B.
For each student, we evaluate English Math, Chinese Math, and
Long-Horizon Math.
Across both model sizes, OPD on Code or Science queries generally
improves mathematical reasoning, even though no Math queries are used
in these runs.
The gains extend beyond standard English benchmarks to Chinese
problems and composed long-horizon problems.
This confirms that the teacher's mathematical capability can be
transferred through student trajectories collected from non-Math
domains.

The complete curves also reinforce the model-origin effect reported in
the main text.
Same-origin teachers typically produce stronger and more stable gains
across the three evaluation distributions.
Within the same teacher lineage, the differences among Math-, Code-,
and Science-supervised runs are comparatively small, whereas changing
the teacher origin produces a larger performance gap.
Thus, the teacher--student post-training relationship can have a
stronger effect on cross-domain transfer than the nominal domain of the
training queries.

\begin{figure*}[t]
    \centering
    \begin{subfigure}[b]{0.99\linewidth}
        \centering
        \includegraphics[width=\linewidth]{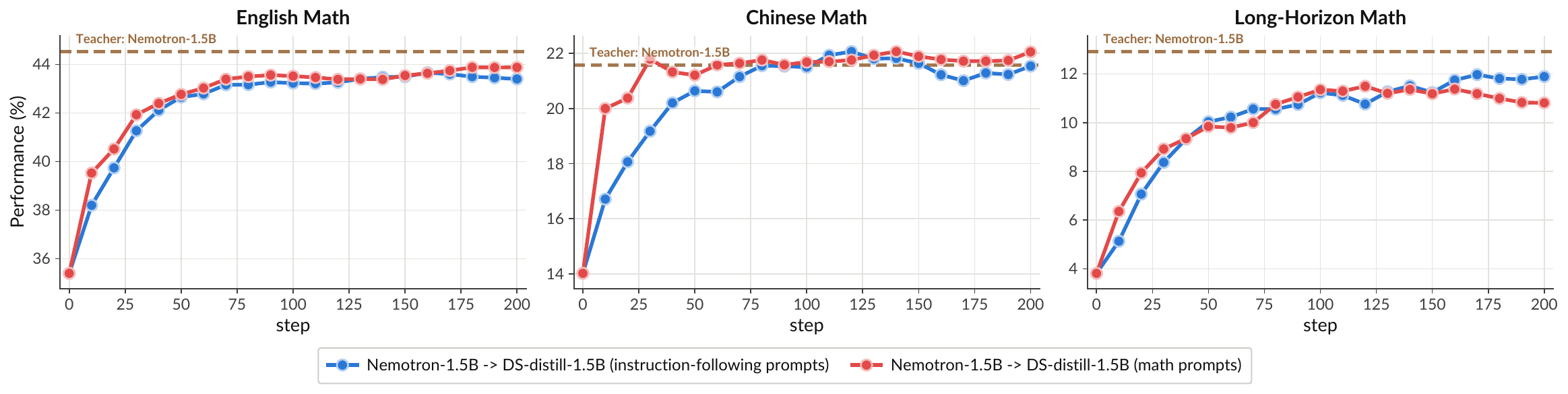}
        \caption{Nemotron-1.5B to DS-distill-1.5B}
        \label{fig:if_to_math_1d5b}
    \end{subfigure}

     \begin{subfigure}[b]{0.99\linewidth}
        \centering
        \includegraphics[width=\linewidth]{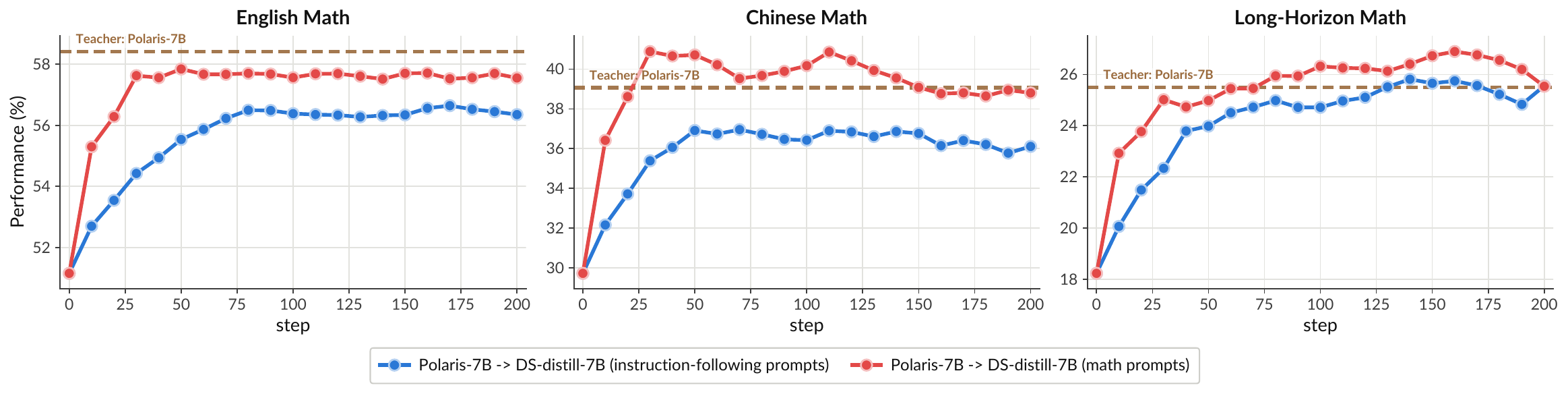}
        \caption{Polaris-7B to DS-distill-7B}
        \label{fig:if_to_math_7b}
    \end{subfigure}
    
    \caption{Training on instruction-following data and generalizing on math-related benchmarks.}
    \label{fig:if_to_math_cross_domain_generalization}
\end{figure*}

We further examine whether instruction-following data can serve as a
carrier for mathematical capability transfer.
Figure~\ref{fig:if_to_math_cross_domain_generalization} reports results
for Nemotron-1.5B $\rightarrow$ DS-distill-1.5B and
Polaris-7B $\rightarrow$ DS-distill-7B.
Both configurations are evaluated on English Math, Chinese Math, and
Even when the student trajectories are collected from
instruction-following prompts, the teacher can still transfer
capabilities that are not explicitly represented by the nominal
training domain.
Together with the previous results, this suggests that the training
queries primarily determine where teacher supervision is elicited,
rather than strictly restricting which teacher capabilities can be
transferred.

\subsection{Full Mathematical Results for MOPD}
\label{sec:appendix_mopd_full_math}

Table~\ref{tab:mopd_math_full_result} reports the complete mathematical
reasoning results of the MOPD experiments at training step 200.
Under the first teacher--domain assignment, JustRL provides Math
supervision and Nemotron provides science/IF supervision.
JustRL-dominant mixtures generally maintain stronger mathematical
performance.
The $25/8$ configuration achieves the highest average score of
$19.6\%$, slightly exceeding both the JustRL-only configuration and the
balanced $1/1$ mixture.
Also, mathematical performance decreases substantially once
Nemotron becomes dominant.
The average falls from $19.1\%$ for the balanced mixture to $17.4\%$
for $2/25$ and $14.9\%$ for the Nemotron-only configuration.
The relationship is not strictly monotonic on every individual
benchmark, but the aggregate trend shows that excessive Nemotron
supervision shifts the student toward Nemotron's weaker mathematical
capability.

The reversed assignment provides complementary evidence.
Here, Nemotron supervises Math queries, whereas JustRL supervises
science/IF queries.
Despite this reassignment, increasing the proportion of JustRL
supervision still improves the student's mathematical performance.
The JustRL-only configuration reaches an average of $17.5\%$, compared
with $15.8\%$ for both the balanced and Nemotron-only configurations.
This result cannot be explained by the nominal training-domain
assignment, since JustRL does not supervise Math data in this setting.
Instead, its stronger mathematical capability is transferred through
science/IF trajectories.
The full benchmark results therefore confirm that domain routing does
not isolate a teacher's influence to its assigned domain.

\begin{table*}[htbp]
\centering
\caption{MOPD performance comparison across five mathematical reasoning benchmarks (training step 200).}
\label{tab:mopd_math_full_result}
\resizebox{\textwidth}{!}{%
\begin{tabular}{c|ccccc|c}
\toprule
\textbf{Configuration} & \textbf{BeyondAIME} & \textbf{OlymMATH-e(en)} & \textbf{OlymMATH-e(zh)} & \textbf{OlymMATH-h} & \textbf{AIME24-n2} & \textbf{Average} \\
\midrule
\midrule
DS-distill-1.5B & $9.6\%$ & $14.2\%$ & $11.2\%$ & $3.2\%$ & $6.0\%$ & $8.8\%$ \\
JustRL-1.5B & $19.0\%$ & $32.6\%$ & $23.0\%$ & $5.7\%$ & $16.3\%$ & $19.3\%$ \\
Nemotron-1.5B & $17.6\%$ & $29.2\%$ & $16.6\%$ & $5.6\%$ & $14.3\%$ & $16.7\%$ \\
\midrule
\midrule
\multicolumn{7}{c}{\textbf{\textit{Math Teacher: JustRL-1.5B, science/IF Teacher: Nemotron-1.5B, Student: DS-distill-1.5B}}} \\
\midrule

$\mathbf{\text{JustRL}/\text{Nemotron}=25/0}$ & $19.0\%$ & $32.6\%$ & $23.0\%$ & $5.7\%$ & $16.3\%$ & $19.3\%${\footnotesize \textcolor{green!60!black}{(\textbf{+0.2 pp})}} \\
$\mathbf{\text{JustRL}/\text{Nemotron}=25/2}$ & $19.4\%$ & $33.3\%$ & $23.0\%$ & $5.9\%$ & $14.1\%$ & $19.1\%${\footnotesize \textcolor{gray!60!black}{(\textbf{+0.0 pp})}} \\
$\mathbf{\text{JustRL}/\text{Nemotron}=25/8}$ & $19.6\%$ & $34.5\%$ & $23.2\%$ & $5.5\%$ & $15.3\%$ & $19.6\%${\footnotesize \textcolor{green!60!black}{(\textbf{+0.5 pp})}} \\
$\mathbf{\text{JustRL}/\text{Nemotron}=1/1}$ & $20.2\%$ & $32.4\%$ & $20.6\%$ & $5.0\%$ & $17.3\%$ & $19.1\%${\footnotesize \textcolor{black!60!black}{(\textbf{baseline})}} \\
$\mathbf{\text{JustRL}/\text{Nemotron}=8/25}$ & $19.7\%$ & $32.5\%$ & $22.7\%$ & $5.5\%$ & $15.7\%$ & $19.2\%${\footnotesize \textcolor{green!60!black}{(\textbf{+0.1 pp})}} \\
$\mathbf{\text{JustRL}/\text{Nemotron}=2/25}$ & $19.2\%$ & $31.0\%$ & $19.2\%$ & $5.2\%$ & $12.3\%$ & $17.4\%${\footnotesize \textcolor{red!60!black}{(\textbf{-1.7 pp})}} \\
$\mathbf{\text{JustRL}/\text{Nemotron}=0/25}$ & $15.9\%$ & $25.3\%$ & $15.7\%$ & $3.7\%$ & $13.8\%$ & $14.9\%${\footnotesize \textcolor{red!60!black}{(\textbf{-4.2 pp})}} \\

\midrule
\midrule
\multicolumn{7}{c}{\textbf{\textit{Math Teacher: Nemotron-1.5B, science/IF Teacher: JustRL-1.5B, Student: DS-distill-1.5B}}} \\
\midrule

$\mathbf{\text{JustRL}/\text{Nemotron}=25/0}$ & $18.2\%$ & $29.8\%$ & $21.1\%$ & $5.8\%$ & $12.3\%$ & $17.5\%${\footnotesize \textcolor{green!60!black}{(\textbf{+1.7 pp})}} \\
$\mathbf{\text{JustRL}/\text{Nemotron}=25/8}$ & $17.4\%$ & $28.0\%$ & $17.8\%$ & $4.5\%$ & $14.8\%$ & $16.5\%${\footnotesize \textcolor{green!60!black}{(\textbf{+0.7 pp})}} \\
$\mathbf{\text{JustRL}/\text{Nemotron}=1/1}$ & $17.3\%$ & $26.6\%$ & $16.8\%$ & $4.4\%$ & $13.9\%$ & $15.8\%${\footnotesize \textcolor{black!60!black}{(\textbf{baseline})}} \\
$\mathbf{\text{JustRL}/\text{Nemotron}=8/25}$  & $17.1\%$ & $26.2\%$ & $16.8\%$ & $4.7\%$ & $14.3\%$ & $15.8\%${\footnotesize \textcolor{gray!60!black}{(\textbf{-0.0 pp})}} \\
$\mathbf{\text{JustRL}/\text{Nemotron}=0/25}$ & $17.4\%$ & $26.5\%$ & $17.0\%$ & $4.5\%$ & $13.6\%$ & $15.8\%${\footnotesize \textcolor{gray!60!black}{(\textbf{-0.0 pp})}} \\
\bottomrule
\end{tabular}%
}
\end{table*}